\documentclass[twoside]{article}

\usepackage{bm}
\usepackage{booktabs}
\usepackage{graphicx}
\usepackage{float}
\usepackage{graphics}
\usepackage[hidelinks]{hyperref}
\newsavebox\mybox
\usepackage{morefloats}
\usepackage{subcaption}
\usepackage{tabularx}
\usepackage{tikz}
\usetikzlibrary{backgrounds}
\usepackage{varioref}
\usepackage{amssymb}
\usepackage{dsfont}

\usepackage{amsthm,amsmath,amsfonts,amssymb}
\usepackage{mathtools}
\usepackage{mymacros}

\usepackage[accepted]{aistats2021}
\usepackage[round]{natbib}

\begin{document}

%
\runningtitle{Deep Generative Pattern-Set Mixture Models for Nonignorable Missingness}

%
\runningauthor{Sahra Ghalebikesabi, Rob Cornish,  Luke J. Kelly,  Chris Holmes}

\twocolumn[

\aistatstitle{Deep Generative Pattern-Set Mixture Models\\ for Nonignorable Missingness}
\addtocounter{footnote}{1}

\aistatsauthor{Sahra Ghalebikesabi \And Rob Cornish \And  Luke J. Kelly  \And  Chris Holmes }

\aistatsaddress{University of Oxford \And University of Oxford  \And CEREMADE, CNRS \\ Université Paris-Dauphine  \And University of Oxford}
]

\begin{abstract}
We propose a variational autoencoder architecture to model both ignorable and nonignorable missing data using pattern-set mixtures as proposed by Little (1993). Our model explicitly learns to cluster the missing data into missingness pattern sets based on the observed data and missingness masks. Underpinning our approach is the assumption that the data distribution under missingness is probabilistically semi-supervised by samples from the observed data distribution. Our setup trades off the characteristics of ignorable and nonignorable missingness and can thus be applied to data of both types. We evaluate our method on a wide range of data sets with different types of missingness and achieve state-of-the-art imputation performance. Our model outperforms many common imputation algorithms, especially when the amount of missing data is high and the missingness mechanism is nonignorable.
\end{abstract}

\section{INTRODUCTION}

Missing data is a ubiquitous problem in real-world applications. In imaging, for instance, inpainting algorithms are used to restore damaged images or to remove selected objects \citep{ber00}. In medical applications, some patient records never get collected because of missed appointments or because information acquisition was too cost intensive. In other cases, some of the data may have simply been lost. Since the 1970s, a rapidly growing body of literature has developed imputation algorithms that appropriately fill in the missing data. 
Missing data is commonly classified into three categories: missing completely at random (MCAR), missing at random (MAR) and missing not at random (MNAR) \citep{rubin1976inference}. When data are MCAR, the missingness is independent of both the observed and missing variables. 
Data are MAR when the missingness depends on the observed variables only. 
And they are said to be MNAR when the missingness mechanism depends not only on the observed variables but also on the missing values. 
Imputing MNAR data bears the highest difficulties as it requires modeling the missing data mechanism. It is thus not sufficiently analyzed in the current machine learning literature. 

In this paper, we exemplary explain some existing deep generative imputation methods in the framework of nonignorable missingness models \citep{rubin1976inference}. We propose a VAE-based imputation algorithm derived from one of these nonignorable missingness models, the pattern-set mixture model, which was introduced by \cite{lit93}. Our model encodes the nonignorable missing data mechanism in a latent variable. To prevent underidentification, we impose probabilistic semi-supervision. This approach prevents overfitting to the observed data. An additional benefit of our method is that it explicitly provides meaningful clusters of the data that can be exploited to cluster the population based on the similarity of their missing data mechanism. We show that the model achieves state-of-the-art performance on a variety of data sets. Furthermore, we provide an implementation of our algorithm and all corresponding experiments online.\footnote{\url{https://github.com/sghalebikesabi/PSMVAE}}

\section{RELATED WORK}
Common imputation approaches include the EM algorithm \citep{dem77}, MICE \citep{buuren2010mice}, matrix completion \citep{mazumder2010spectral} and MissForest \citep{ste12}. With the recent development of deep generative models, there has been an increase in generative imputation algorithms. Deep generative models such as generative adversarial nets (GANs) \citep{goodfellow2014generative} or variational autoencoders (VAEs) \citep{kin13, rezende2014stochastic} have been used for missingness imputation since their introduction. 

One of the most popular examples is GAIN \citep{yoo18}. It uses a generative adversarial network in which the generator takes the observed data vector as an input and outputs an imputed data vector. The discriminator determines which variables were actually observed based on partial information of the true missingness mask. Another GAN-based imputation algorithm, MisGAN \citep{li2018learning}, learns one generator and one discriminator solely for the missingness mask, and simultaneously learns a separate generator and discriminator for the data and the missingness mask.

\cite{nazabal2020handling} introduce an objective function in order to directly train VAEs with a Gaussian Mixture prior on data with missingness. 
MIWAE \citep{mat18} is an importance-weighted autoencoder that achieves state-of-the-art imputation performance by maximizing a tight lower bound of the log-likelihood of the observed data.

All these imputation algorithms have in common that they do not incorporate assumptions on the missingness mechanisms and thus cannot be applied when the data is MNAR. To tackle this problem, \cite{ipsen2020not} introduced an extension of MIWAE, the not-MIWAE, that handles MNAR data by explicitly modeling the conditional distribution of the missingness mask given the data. 
Such an approach is, however, not robust to the misspecification of the missingness mechanism. \cite{collier2020vaes} develop a latent variable  model  that considers the observed data to be the result of a corruption process which depends on a binary missingness mask and can be applied on data with ignorable and nonignorable missingness. Nonetheless, the model architecture has to be chosen dependent on the underlying missingness mechanism. Recently, \cite{sportisse2019estimation} have shown the identifiability of the parameters of probabilistic principal components analysis for a class of MNAR mechanisms. 
Missingness not at random has also been tackled using discriminative approaches such as matrix completion in the recommender system literature \citep{wang2019doubly}. 

\section{PROBLEM FORMULATION}
Let $\mathbf{x}=(x_1, ..., x_d)$ be a random variable taking values in the $d$-dimensional space $\mathcal{X}=\mathcal{X}_1\times...\times \mathcal{X}_d$. We further assume that the missingness mask $\mathbf{m}$ is a random variable defined on $\{0, 1\}^{d}$. The missingness mask is defined such that $x_j$ is observed for $m_j=1$, and missing otherwise. We denote the joint distribution of $\mathbf{x}$ and $\mathbf{m}$ by $P_{\theta}(\mathbf{x}, \mathbf{m})$. Following \cite{yoo18}, we also introduce a random variable $\mathbf{x_{obs}}=(x_{obs, 1}, ..., x_{obs, d})$ which takes values in $\mathcal{X}_{obs}=(\mathcal{X}_1\cup\{*\})\times\cdots\times(\mathcal{X}_d\cup\{*\})$ where $*$ is a point not in $\mathcal{X}_1\cup....\cup\mathcal{X}_d$ and represents unobserved data points. The random variable $\mathbf{x_{obs}}$ is then defined by
\begin{align*}
    x_{obs, j} = 
    \begin{cases}
        x_{j}, & \text{if }m_j=1\\
        *, & \text{otherwise.}
    \end{cases}
\end{align*}
Building upon this, we introduce another random variable $\mathbf{x_{mis}}$ with $x_{mis, j} = x_{j}$ if $m_j=0$ and
$x_{mis, j} = *$ otherwise.
We can now retrieve $\mathbf{x}$ as 
\begin{align}
\mathbf{x} = \mathbf{m} \odot \mathbf{x_{obs}} + (\mathbf{1}-\mathbf{m}) \odot \mathbf{x_{mis}}, \label{x}
\end{align}
where $\odot$ denotes the Hadamard product.
Note that both $\mathbf{x_{mis}}$ and $\mathbf{x_{obs}}$ are defined to be $d$-dimensional random variables. We can interpret such an approach by imagining self-masked missingness, which means that the missingness probability of each covariate only depends on the covariate itself such that $P_{\theta}(m_j|\mathbf{x})=P_{\theta}(m_j|x_j)$. Then ${x_{mis, j}}$ denotes the outcome of that covariate under the treatment $m_j=0$ and ${x_{obs, j}}$ the outcome under the treatment $m_j=1$. When we want to refer to only the observed or missing observations, we instead write $\mathbf{x'_{obs}}=\{x_j\mid m_j=1\text{ for }j\in\{1,...,d\}\}$ and  $\mathbf{x'_{mis}}=\{x_j\mid m_j=0\text{ for }j\in\{1,...,d\}\}$.

In the imputation setting, we assume that we are given $n$ i.i.d. copies $\mathbf{x_{obs}^1},...,\mathbf{x_{obs}^n}$ of $\mathbf{x_{obs}}$. From these observations, we can infer the missingness masks $\mathbf{m^1},....,\mathbf{m^n}$ as realizations of $\mathbf{m}$. We write the observed data as $\mathcal{D}=\{\mathbf{x_{obs}^1},...,\mathbf{x_{obs}^n}, \mathbf{m^1},....,\mathbf{m^n}\}$. When we impute the missing data, we aim to recover 
$P_{\theta}(\mathbf{x_{mis}}|\mathbf{x_{obs}}, \mathbf{m})$. 
\section{NONIGNORABLE MISSINGNESS MODELS}
When data are MCAR (i.e. $P_{\theta}(\mathbf{x}, \mathbf{m}) = P_{\theta}(\mathbf{x})P_{\theta}(\mathbf{m})$) or MAR (i.e. $P_{\theta}(\mathbf{x}, \mathbf{m})=P_{\theta}(\mathbf{x}) P_{\theta}(\mathbf{m}|\mathbf{x_{obs}'})$), we can maximize the data likelihood without modelling the missing-data mechanism since
\begin{align*}
    P_{\theta}(\mathbf{x'_{obs}}, \mathbf{m}) =& \int P_{\theta}(\mathbf{x'_{obs}}, \mathbf{x'_{mis}})P_{\theta}(\mathbf{m}|\mathbf{x'_{obs}}, \mathbf{x'_{mis}}) d\mathbf{x'_{mis}}\\
    =& P_{\theta}(\mathbf{x'_{obs}})P_{\theta}(\mathbf{m}|\mathbf{x'_{obs}}).
\end{align*}
When data are MNAR, the missing-data mechanism is nonignorable and has to be modeled within a maximum likelihood framework. \citet{lit19} differentiate three ways of modelling the joint distribution of $\mathbf{x}$ and $\mathbf{m}$ in this case. 

\emph{Selection models} factorize the joint distribution as
$ P_{\theta}(\mathbf{x}, \mathbf{m}) = P_{\theta}(\mathbf{x}) P_{\theta}(\mathbf{m}|\mathbf{x}).$
This factorization goes along with intuitive missing-data mechanisms: In the MNAR case, the complete data $x$ can be seen as the cause why some variables are missing. 
Not-MIWAE \citep{ipsen2020not} can be categorized as a selection model that uses MIWAE to model $P_{\theta}(\mathbf{x})$ and an additional layer (through a neural network layer or a logistic regression) on top of $P_{\theta}(\mathbf{x})$ to model $P_{\theta}(\mathbf{m}|\mathbf{x})$. GAIN \citep{yoo18} makes an MCAR assumption in the setting of a selection model. As such it imputes the missing values by modeling the data distribution $P_{\theta}(\mathbf{x})$ using a neural network with one hidden layer, and then estimates $P_{\theta}(\mathbf{m}|\mathbf{x}, \mathbf{h})$ where $\mathbf{h}$ is a hint variable that indicates what $\mathbf{m}$ looks like.

\emph{Pattern mixture models} factorize the joint distribution as
$
P_{\theta}(\mathbf{x},\mathbf{m})= P_{\theta}(\mathbf{x}|\mathbf{m}) P_{\theta}(\mathbf{m}),
$
where $P_{\theta}(\mathbf{m})$ is a categorical distribution and $P_{\theta}(\mathbf{x},\mathbf{m})$ is as a result a mixture of distributions.
The drawback of such a parameterization is that $P_{\theta}(\mathbf{x}|\mathbf{m})$ is conditional on a high-dimensional categorical variable whose categories are often not completely observed which leads to the distribution of the missing data being underidentified without any additional assumptions \citep{lit93}. Despite this, pattern mixture models are applied when there is an interest in $P_{\theta}(\mathbf{x}|\mathbf{m})$. 
For instance, a drug company might be interested in predicting the lab values of patients that dropped out of a clinical trial. 
\cite{collier2020vaes} propose a latent variable model that optimizes the lower bound of ${P_{\theta}(\mathbf{x_{obs}}|\mathbf{m})}$ and as such constitutes a special type of pattern mixture model.

In order to combine the benefits of both factorizations, \cite{lit93} introduced \emph{pattern-set mixture models}. These models have an additional latent variable $\mathbf{r}$ with realizations in $\{1,...,k\}$ that clusters the missingness patterns into $k$ missingness pattern-sets. In each missingness pattern-set, the missing data mechanism is modeled using a selection model. The joint distribution can then be written as 
\begin{align}\label{eq:psm} 
 P_{\theta}(\mathbf{x},\mathbf{m}, \mathbf{r}) 
= P_{\theta}(\mathbf{r})&P_{\theta}(\mathbf{x}|\mathbf{r}) P_{\theta}(\mathbf{m}|\mathbf{x},\mathbf{r}).
\end{align}
For $k=1$ this model reduces to a selection model. Compared to pattern mixture models, it requires fewer parameters (because we are borrowing strengths across the clusters), is less prone to underidentification and has thus more statistical power. Regardless of this, it still allows us to cluster the population into interesting categories. In a clinical trial, there might be some patients that dropped out for reasons unrelated to the study, while other patients dropped out because of adverse reactions to the treatment. The objective is that the missingness pattern-set $\mathbf{r}$ corresponds to these different missingness mechanisms. It summarizes similar missingness patterns and thus simplifies downstream tasks. The HIVAE model \citep{nazabal2020handling}, a Gaussian Mixture VAE, can be seen as a pattern-set mixture model for data that are MAR and where $P_{\theta}(\mathbf{m}|\mathbf{x},\mathbf{r})$ is ignored in the maximum likelihood inference. 

\cite{wu1988estimation} propose \textit{shared-parameter models} which build upon the assumption that there exists a latent random variable $\mathbf{z}\in\mathbb{R}^b$ for some $b<n$ conditional on which the missing model and the data model are independent:
$
     P_{\theta}(\mathbf{x},\mathbf{m}|\mathbf{z})= 
     P_{\theta}(\mathbf{x}|\mathbf{z}) P_{\theta}(\mathbf{m}|\mathbf{z}).
$

These specifications are equivalent only when the data is MCAR \citep{lit19}. The choice of the model should be made based on the underlying data problem. We argue that in the case of high-dimensional data sets, such as images, where machine learning algorithms proved to be more useful than classical statistical approaches, pattern mixture models are not feasible considering the high-dimensional space of $\mathbf{m}$. As selection models fall within the class of pattern-set mixture models, we use a combination of pattern-set mixture models and shared-parameter models for building a deep generative pattern-set mixture model.

\section{DEEP GENERATIVE PATTERN-SET MIXTURE MODEL}

We now introduce an imputation approach that combines ideas of variational autoencoders and pattern-set mixture models. In contrast to other machine learning imputation methods such as HIVAE \citep{nazabal2020handling} or MIWAE \citep{mat18}, we thus aim to model the joint distribution $P_{\theta}(\mathbf{x}, \mathbf{m})$ instead of the marginal $P_{\theta}(\mathbf{x})$.

\subsection{Generative Model}
We will now define a generative model for $P_{\theta}(\mathbf{x}, \mathbf{m})$. More specifically, we will model $P_{\theta}(\mathbf{x_{obs}}, \mathbf{m}, \mathbf{x_{mis}})$ where we assume for now that $\mathbf{x_{mis}}$ is a latent variable. Then, $P_{\theta}(\mathbf{x}, \mathbf{m})$ follows from Equation~\ref{x}. Assuming the pattern-set mixture model holds, we introduce an additional latent categorical variable $\mathbf{r}$ which groups the missingness patterns into sets. Following Equation~\ref{eq:psm} and assuming that $P_{\theta}(\mathbf{m}|\mathbf{x_{obs}}, \mathbf{x_{mis}}, \mathbf{r})=P_{\theta}(\mathbf{m}|\mathbf{x}, \mathbf{r})$, we can now write the joint distribution as
\begin{align*}
   &{P_{\theta}(\mathbf{x_{obs}}, \mathbf{m}, \mathbf{x_{mis}}, \mathbf{r})} \\
   &\quad= P_{\theta}(\mathbf{r})P_{\theta}(\mathbf{x_{obs}}, \mathbf{x_{mis}}|\mathbf{r}){P_{\theta}(\mathbf{m}|\mathbf{x}, \mathbf{r})  }.
\end{align*}
Since $\mathbf{r}$ is a categorical variable that only captures the pattern-set of an observation, we introduce an additional continuous latent variable $\mathbf{z}$ that models the latent interaction of $\mathbf{x_{mis}}$ and $\mathbf{x_{obs}}$. Given $\mathbf{z}$ and $\mathbf{r}$, we then assume the joint of $\mathbf{x_{mis}}$ and $\mathbf{x_{obs}}$ to fully factorize implying $$P_{\theta}(\mathbf{x_{mis}}, \mathbf{x_{obs}})=\prod_{j=1}^d P_{\theta}({x_{mis, j}}|\mathbf{z},\mathbf{r})P_{\theta}({x_{obs,j}}|\mathbf{z},\mathbf{r}).$$
If additional information on the missing data mechanism $P_{\theta}(\mathbf{m}|\mathbf{x}, \mathbf{r})$ is available, we can write the generative model as
\begin{align}
    &P_{\theta}(\mathbf{x_{obs}}, \mathbf{m}, \mathbf{x_{mis}}, \mathbf{z}, \mathbf{r})  \label{pmxr} \\
    & =P_{\theta}(\mathbf{m}|{\mathbf{x}}, \mathbf{r}) P_{\theta}(\mathbf{x_{obs}}| \mathbf{r}, \mathbf{z}) P_{\theta}({\mathbf{x_{mis}}}|\mathbf{r}, \mathbf{z})P_{\theta}(\mathbf{z}|\mathbf{r})P_{\theta}(\mathbf{r}), \nonumber
\end{align}
as is visualized in Figure~\ref{fig:pmxr}.
%
While \cite{ipsen2020not} only show how one uniform missing model can be parameterized for the whole population, our approach allows to account for different missingness models. This can be interesting when part of the data is assumed to be MCAR while some observations can be MNAR.

We parameterize the generative model of the VAE using a neural network for improved data fit. Allowing the missing model $P_{\theta}(\mathbf{m}|\mathbf{x}, \mathbf{r})$ to be parameterized by a neural network has the disadvantage that the fit of $P_{\theta}(\mathbf{x_{obs}}, \mathbf{x_{mis}}|\mathbf{r})$ suffers from the flexibility of the missing model. This statement is strengthened by the empirical results of \cite{ipsen2020not}. Since for any MNAR model there is an MAR model with equal fit to the observed data, it is not possible to test for MNAR without any further assumptions on the missing data mechanism \citep{molenberghs2008every}. For this reason, it is important to choose an imputation algorithm that finds a trade off between flexibility of the missingness model and the distortion it induces into the data model when the data are MCAR. We find that this dilemma can be solved by the assumption made when using shared parameter models that $\mathbf{x_{obs}}, \mathbf{x_{mis}}$ and $\mathbf{m}$ are independent conditional on the pattern-set indicator $\mathbf{r}$ and the continuous latent representation $\mathbf{z}$. Such a model has been proven to be more robust to model specification \citep{harel2009partial}.
When no additional information on $P_{\theta}(\mathbf{m}|\mathbf{x})$ is available, we thus formalize the generative model as
\begin{align}
    \label{eq:gen}
    &P_{\theta}(\mathbf{x_{obs}}, \mathbf{m}, \mathbf{x_{mis}},  \mathbf{z}, \mathbf{r}) \\
    &\quad= P_{\theta}(\mathbf{r}) P_{\theta}(\mathbf{z}|\mathbf{r}) P_{\theta}(\mathbf{x_{mis}}|\mathbf{r}, \mathbf{z}) P_{\theta}(\mathbf{x_{obs}}|\mathbf{r}, \mathbf{z}) P_{\theta}(\mathbf{m}|\mathbf{r}, \mathbf{z}), \nonumber 
\end{align}
as illustrated in Figure~\ref{fig:pmz}.
The missing values are imputed by sampling from $P_{\theta}(\mathbf{x_{mis}}|\mathbf{x_{obs}}, \mathbf{m}) = P_{\theta}(\mathbf{x_{obs}}|\mathbf{r}, \mathbf{z})$ which follows from the conditional independence of $\mathbf{x_{obs}}, \mathbf{m}$ and $\mathbf{x_{mis}}$ given $\mathbf{r}$ and $\mathbf{z}$. 

\subsection{Probabilistic Semi-Supervision}
We use likelihood-based variational inference to learn the joint distribution $P_{\theta}(\mathbf{x}, \mathbf{m})$. As we do not observe the missing data, we can however only optimize for the parameters of $P_{\theta}(\mathbf{x_{obs}}, \mathbf{m})$. A sensible choice is to treat $\mathbf{x_{mis}}$ as a latent factor learned from $\mathbf{x_{obs}}$ and $\mathbf{m}$ \citep{nazabal2020handling}. 
Without any further assumptions, the distribution of $\mathbf{x_{mis}}$ would be underidentified given the observed data as, among other reasons, any orthogonal transformation of $\mathbf{x_{mis}}$ would not affect the observed data likelihood $P_{\theta}(\mathbf{x_{obs}}, \mathbf{m})$ \citep{khemakhem2020variational}. 


%
We can regularize the problem and introduce smoothness through a novel process involving information sharing and semi-supervision \citep{kin14, joy2020rethinking}. For any covariate of any sample ${x_{obs, j}^i}$ with $j\in\{1,...,d\}$ and $i\in\{1,...,n\}$, we assume ${x_{obs, j}^i}$ is not only an observation of $x_{obs, j}$, but also of ${x_{mis, j}}$ with probability $1-\pi_{i, j}$ if $m_{i,j}=1$.
Put simply, we sample each ${x^i_{mis, j}}$ from
$$
     y_{i,j} P_{\theta}(\widetilde{x}_{mis,j}) + (1-y_{i,j})\mathds{1}(x^i_{obs, j}), 
$$
where $\mathds{1}$ is the indicator function, $\widetilde{x}_{mis, j}$ is a latent auxiliary variable that describes the unobserved dynamics of the missing data, and $y_j$ is an independent Bernoulli random variable with known success probability $\pi'$ if $m_j=1$, and with probability 1 otherwise. We then define the augmented data set as
\begin{align}
    \mathcal{D}_{\bm{\pi}}(\mathbf{y^1}, ..., \mathbf{y^n})\vcentcolon=\mathcal{D}\cup \{x_{mis, j}^i; y_{j}^i=0\}_{i, j}. \label{eq:dataset}
\end{align}
For simplicity, let us assume for now that we observe a single univariate data point $x^0_{obs, 0}$ with $m^0_{0}=0$ such that $\mathcal{D}=\{x^0_{obs, 0}, m^0_{0}\}$. With probability $1-\pi'$, it holds that $y_0^0=0$. We then assume that $x^0_{mis, 0}$ is also observed with value $x^0_{obs, 0}$ and the augmented data set is thus $\mathcal{D}_{\pi}(y_0^0=0)=\{x^0_{obs, 0}, m^0_{0}, x^0_{mis, 0}\}$. In this case, we maximize the likelihood $P_{\theta}(x_{obs, 0}, m_{0}, x_{mis, 0}|\mathcal{D}_{\pi}(y_0^0=0))$. 
With probability $\pi'$, however, it holds that $y_0^0=1$ and $x^0_{mis, 0}$ is assumed to be unobserved. We then have $\mathcal{D}_{\pi}(y_0^0=1)=\{x^0_{obs, 0}, m^0_{0}\}=\mathcal{D}$. We now maximize the likelihood $P_{\theta}(x_{obs, 0}, m_{0}|\mathcal{D}_{\pi}(y_0^0=1))$. Since we know the true distribution of $y^0_0$, we can also marginalize out $y_0$ and maximize the weighted likelihood \begin{align*}
    &\pi' P_{\theta}(x_{obs, 0}, m_{0}|\mathcal{D}_{\pi}(y_0^0=1)) \\
    &\quad + (1-\pi') P_{\theta}(x_{obs, 0}, m_{0}, x_{mis, 0}|\mathcal{D}_{\pi}(y_0^0=0)).
\end{align*}

In a more general setting, we can write the expected likelihood given the augmented data set as
\begin{align*}
    &\mathbb{E}_{\mathbf{y}}[P_{\theta}(\mathbf{x_{obs}}, \mathbf{x_{mis, 1-y}}, \mathbf{m}|\mathcal{D}_{\bm{\pi}}(\mathbf{y}))] \\
    &\quad={\bm{\pi}} P_{\theta}(\mathbf{x_{obs}}, \mathbf{m}|\mathcal{D}_{\bm{\pi}}(\mathbf{1})) \\
    &\qquad+ (\mathbf{1}- \bm{\pi})P_{\theta}(\mathbf{x_{obs}}, \mathbf{m}, \mathbf{x_{mis}}|\mathcal{D}_{\bm{\pi}}(\mathbf{0})),
\end{align*}
where $\mathbf{x_{mis, 1-y}}\vcentcolon=\{x_{mis, j}| y_j=0\text{ for } j\in\{1,...,d\} \}$, and $\mathbf{1}$ and $\mathbf{0}$ are $d$-vectors of ones and zeros respectively.
We only assume semi-supervision for the covariates $x_{mis,j}|(m_j=1)\in\{*\}$ which drop out in the generation process of $x_j$ \eqref{x}. The parameter $\pi$ can thus be interpreted as confidence on the ignorability of the missing model: the greater $\pi$ is, the less likely $x_{mis, j}$ stems from an observed distribution. Note that this approach is equivalent to a biased data augmentation approach and that we do not modify the generative model here. As proven in the supplements, it holds:

\textbf{Proposition 1.} \textit{
Assume that we observe data $\mathcal{D}$ sampled from one of the generative models defined according to \eqref{pmxr} or \eqref{eq:gen}. For $\pi'<1$, it holds that the distributional parameter ${\mu}$ of $\mathbf{x_{mis}}|\mathbb{E}_{\mathbf{y}}(\mathcal{D}_{\pi}(\mathbf{y})))\sim\mathcal{N}({\mu}, c)$ for some fixed constant $c$ and $\mathcal{D}_{\pi}(\mathbf{y})$ as defined in \eqref{eq:dataset} is identifiable under the maximum likelihood estimation method.}

Without any additional model assumptions such as the specification of the missing model, the distribution of $\mathbf{\widetilde{x}_{mis}}$ will be underidentified and the estimator of $\mathbf{{x}_{mis}}$ will be biased. In that case, the question arises why we should not just assume $P_{\theta}(\mathbf{x_{mis}})=P_{\theta}(\mathbf{x_{obs}})$ from the beginning. We argue that our unconventional approach to model both $\mathbf{x_{obs}}$ and $\mathbf{x_{mis}}$ as $d$-dimensional vectors and include such a semi-supervised approach introduces smoothing and prevents overfitting to the observed data distribution which is especially beneficial when missingness is high. A similar methodology has been proposed by \cite{szegedy2016rethinking} under the term label-smoothing regularization. 
Contrary to our approach, \cite{nazabal2020handling} assume the union of $\mathbf{x_{obs}}$ and $\mathbf{x_{mis}}$ to be $d$-dimensional. In other words, they assume that only the missing components of $\mathbf{x}$ are latent. As a result the missing data imputation of the HIVAE model can be seen as a special case of our model for $\pi'=0$.
\subsection{Recognition Model}
As already noted, the generative model is learned by maximum likelihood inference. The marginal likelihood of the observed variables, $P_{\theta}(\mathbf{x_{obs}}, \mathbf{m}, \mathbf{x_{mis, 1-y}})$,
is intractable but we can follow \cite{kin13} and define a recognition model $Q_{\phi}$ for modelling the unobserved latent variables $\mathbf{x_{mis, y}}, \mathbf{z}$ and $\mathbf{r}$ given the observed observations $\mathbf{x_{obs}}, \mathbf{m}$ and $\mathbf{x_{mis, 1-y}}$ assuming a simple parametric form:
\begin{align*}
    Q' \vcentcolon= &Q_{\phi}(\mathbf{x_{mis, y}}, \mathbf{z}, \mathbf{r}|\mathbf{x_{obs}}, \mathbf{m}, \mathbf{x_{mis, 1-y}}) \\
    = &Q_{\phi}(\mathbf{r}|\mathbf{x_{obs}}, \mathbf{m}) Q_{\phi}(\mathbf{z}| \mathbf{r}, \mathbf{x_{obs}}, \mathbf{m}, \mathbf{x_{mis, 1-y}}) \\ &\quad \;\; \times Q_{\phi}(\mathbf{x_{mis, y}}|\mathbf{z}, \mathbf{r}, \mathbf{x_{mis, 1-y}}, \mathbf{x_{obs}}, \mathbf{m}). 
\end{align*}
Using Jensen's equality it follows that 
\begin{align*}
    &\log P_{\theta}(\mathbf{x_{obs}}, \mathbf{m}, \mathbf{x_{mis, 1-y}}) \\
    &\qquad\geq \mathbb{E}_{ Q'}[-\log Q' + \log P_{\theta}(\mathbf{x_{obs}}, \mathbf{m}, \mathbf{x_{mis}},  \mathbf{z}, \mathbf{r})]. 
\end{align*}
We can thus fit the model by maximizing this evidence lower bound (ELBO).

\begin{table*}[ht!]
\centering
\caption{ELBO of the proposed model.}
\begin{tabular}{l}
\hline
\hline
\vspace{2mm}
$\begin{aligned}[t] 
\\
    \mathcal{L}(\mathbf{x_{obs}}, \mathbf{m}) &= \mathbb{E}_{Q_{\phi}(\mathbf{r}, \mathbf{z}\mid\mathbf{x_{obs}}, \mathbf{m})} \bigg[
    \log P_{\theta}(\mathbf{x_{obs}}\mid \mathbf{r}, \mathbf{z}) +
     \log \frac{P_{\theta}(\mathbf{z}\mid\mathbf{r})}{ Q_{\phi}(\mathbf{z}\mid\mathbf{r},\mathbf{x_{obs}}, \mathbf{m})}
    +\log \frac{P_{\theta}(\mathbf{r})}{Q_{\phi}(\mathbf{r}\mid \mathbf{x_{obs}}, \mathbf{m})
}\bigg] \\
    &\qquad +
    \mathbf{\pi} \cdot \mathbb{E}_{Q_{\phi}(\mathbf{r}, \mathbf{z}, \mathbf{x_{mis}}\mid\mathbf{x_{obs}}, \mathbf{m})} \bigg[ \log {P_{\theta}(\mathbf{m}\mid \mathbf{r}, \mathbf{z}, \mathbf{x_{obs}}, \mathbf{x_{mis}})} + \log \frac{P_{\theta}(\mathbf{x_{mis}}\mid \mathbf{r}, \mathbf{z})}{Q_{\phi}(\mathbf{x_{mis}}\mid \mathbf{x_{obs}}, \mathbf{m}, \mathbf{z}, \mathbf{r})} \bigg] \\
    &\qquad+  (1- \mathbf{\pi}) \cdot \mathbb{E}_{Q_{\phi}(\mathbf{r}, \mathbf{z}\mid\mathbf{x_{obs}}, \mathbf{m}, \mathbf{x_{mis}})} \bigg[ \log {P_{\theta}(\mathbf{m}\mid \mathbf{r}, \mathbf{z}, \mathbf{x_{obs}}, \mathbf{x_{mis}})} + \log P_{\theta}(\mathbf{x_{mis}}\mid \mathbf{r}, \mathbf{z})   \bigg]
\end{aligned}$ \\ \hline
\end{tabular}
\label{tab1}
\end{table*}
Since the data sampled for $\mathbf{x_{mis, 1-y}}$ is not more informative than $\mathbf{x_{obs}}$ and $\mathbf{y}$,  and $\mathbf{y}$ is defined independent of the missing data mechanism, we will assume in the following that $Q_{\phi}(\cdot|\mathbf{x_{obs}}, \mathbf{m}, \mathbf{x_{mis, 1-y}})=Q_{\phi}(\cdot|\mathbf{x_{obs}}, \mathbf{m})$. The recognition model can be seen in Figure~\ref{fig:infmodel}. Please refer to Table~\ref{tab1} for the ELBO that results from integrating out $\mathbf{y}$. 
This result is proved in the supplementary material. 

Since we parameterize the recognition model $Q_{\phi}(\mathbf{x_{mis, y}}, \mathbf{z}, \mathbf{r}, |\mathbf{x_{obs}}, \mathbf{m})$ as a neural network, the input has to be of a fixed size. We thus implement an input-dropout layer \citep{nazabal2020handling} which takes all data (missing or observed) as input and drops out all the missing observations. The mechanics of this approach are the same as mean imputing the standardized input before applying the VAE model. 

\savebox{\mybox}{
        \begin{tikzpicture}[>=stealth, shorten >=1pt, auto, node distance=2.5cm, scale=0.7, 
            transform shape, align=center, 
            state/.style={circle, draw, minimum size=1.2cm}]
            \node[rounded corners=8pt, draw, fill=gray!25, minimum size=1.2cm] (xobs) at (1,5) {$\mathbf{x_{obs}}, \mathbf{m}$};
            \node[state] (r) at (1,3) {$\mathbf{r}$};
            \node[state] (z) at (1,1) {$\mathbf{z}$};
            \node[state, dashed, label=below:{ for $j\in\{1,...,d\}$}] (xmis) at (1,-1)  {${{x_{mis, j}}}$};

            \path [->] (xobs) edge node[left] { } (r);
            \path [->] (xobs) edge[bend right] node[left] { } (z);
            \path [->] (r) edge node[left] { } (z);
            \path [->, dashed](xobs) edge[bend right] node[left] { } (xmis);
            \path [->, dashed](r) edge[bend left] node[left] { } (xmis);
            \path [->, dashed](z) edge node[left] { } (xmis);
        \end{tikzpicture}

}
\begin{figure}[ht!]
    \centering
        \begin{subfigure}[t]{0.14\textwidth}
        \centering
        \vbox to \ht\mybox{%
        \vfill
        \begin{tikzpicture}[>=stealth, shorten >=1pt, auto, node distance=2.5cm, scale=0.7, 
            transform shape, align=center, 
            state/.style={circle, draw, minimum size=1.2cm}]
            \node[state] (r) at (2,4) {$\mathbf{r}$};
            \node[state] (z) at (2,2) {$\mathbf{z}$};
            \node[state] (xmisgen) at (3,0) {$\mathbf{x_{mis}}$} ;
            \begin{scope}[on background layer]
                \fill[fill=gray!25] (xmisgen.225) arc [start angle=225, end angle=405, radius=6mm];
            \end{scope}
            \node[state, fill=gray!25] (xobs) at (1,0) {$\mathbf{x_{obs}}$} ;
            \node[state] (m) at (2,-2) {$\mathbf{m}$} ;

            \path [->](r) edge node { } (z);
            \path [->](r) edge[bend left] node { } (xmisgen);
            \path [->](r) edge[bend right] node { } (xobs);
            \path [->](r) edge[bend left=60] node { } (m);
            \path [->](xobs) edge  node { } (m);
            \path [->](xmisgen) edge node { } (m);
            \path [->](z) edge node { } (xobs);
            \path [->](z) edge node { } (xmisgen);
        \end{tikzpicture}
        \vfill
        }
        \caption{Generative model with $P_{\theta}(\mathbf{m}|\mathbf{x},\mathbf{r})$}
        \label{fig:pmxr}
    \end{subfigure}%
\hfill
    \begin{subfigure}[t]{0.14\textwidth}
        \centering
        \vbox to \ht\mybox{%
        \vfill
        \begin{tikzpicture}[>=stealth, shorten >=1pt, auto, node distance=2.5cm, scale=0.7, 
            transform shape, align=center, 
            state/.style={circle, draw, minimum size=1.2cm}]
            \node[state] (r) at (2,4) {$\mathbf{r}$};
            \node[state] (z) at (2,2) {$\mathbf{z}$};
            \node[state] (xmisgen) at (3,0) {$\mathbf{x_{mis}}$} ;
            \begin{scope}[on background layer]
                \fill[fill=gray!25] (xmisgen.225) arc [start angle=225, end angle=405, radius=6mm];
            \end{scope}
            \node[rounded corners=8pt, draw, fill=gray!25, minimum size=1.2cm] (xobsgen) at (1,0) {$\mathbf{x_{obs}}, \mathbf{m}$} ;

            \path [->](r) edge node { } (z);
            \path [->](r) edge[bend left] node { } (xmisgen);
            \path [->](r) edge[bend right] node { } (xobsgen);
            \path [->](z) edge node { } (xobsgen);
            \path [->](z) edge node { } (xmisgen);
        \end{tikzpicture}
        \vfill
        }
        \caption{Generative model with $P_{\theta}(\mathbf{m}|\mathbf{z},\mathbf{r})$}
        \label{fig:pmz}
    \end{subfigure}%
\hfill
    \begin{subfigure}[t]{0.10\textwidth}
        \centering
        \usebox{\mybox}
        \caption{Recognition model.}
        \label{fig:infmodel}
    \end{subfigure}
    \caption{The architecture of the proposed models. Dashed lines and nodes exist with probability $\pi_j=1$ if $m_j=0$ and probability $\pi'$ if $m_j=1$}
    \label{fig:model}
\end{figure}
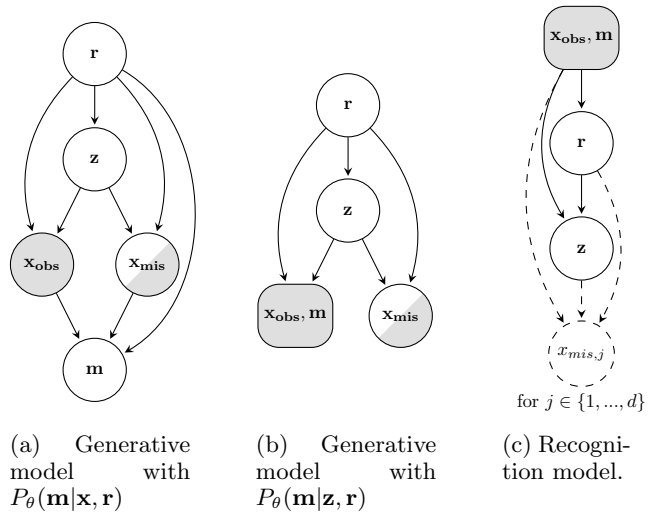
\subsection{Imputation}
The VAE model defined above allows us to learn the predictive distribution of $P_{\theta}(\mathbf{x_{mis}}|\mathbf{x_{obs}}, \mathbf{m})$. When a single value is used for imputing the missing data, we speak of single imputation. For continuous data and when the $l_2$ norm is a relevant error metric, we follow \cite{burda2015importance} and \cite{mat18} and impute the missing data by estimating $\mathbb{E}(\mathbf{x_{mis}}|\mathbf{x_{obs}}, \mathbf{m})$ using importance sampling. This procedure will reduce the noise in our estimation compared to sampling a single value from $P_{\theta}(\mathbf{x_{mis}}|\mathbf{z}, \mathbf{r})$. The above expectation can also be written as 
\begin{align*}
    &\mathbb{E}(h(\mathbf{x_{mis}})|\mathbf{x_{obs}}, \mathbf{m}) \\
    &\quad= \int h(\mathbf{x_{mis}})P_{\theta}(\mathbf{x_{mis}}|\mathbf{x_{obs}}, \mathbf{m}) d\mathbf{x_{mis}} \\
    &\quad= \int h(\mathbf{x_{mis}})P_{\theta}(\mathbf{x_{mis}}|\mathbf{z}) P_{\theta}(\mathbf{z}| \mathbf{x_{obs}}, \mathbf{m})\, d\mathbf{z}\,\,d\mathbf{x_{mis}},
\end{align*}
where $h$ is an absolutely integrable function of $\mathbf{x_{mis}}$ and equal to the identity function in the case of single imputation. 
We use a self-normalized importance weighted estimator with the importance distribution $P_{\theta}(\mathbf{x_{mis}}|\mathbf{z}) Q_{\phi}(\mathbf{z}| \mathbf{x_{obs}}, \mathbf{m})$ and the resulting estimator
\begin{align*}
    \mathbb{E}(h(\mathbf{x_{mis}})|\mathbf{x_{obs}}, \mathbf{m}) \approx  \frac{1}{\sum_{l=1}^L w^{(l)}}\sum_{l=1}^L w^{(l)} h(\mathbf{x_{mis}^{(l)}}),
\end{align*}
where $(\mathbf{x_{mis}}^{(l)}, \textbf{z}^{(l)})_{l=1}^L$ are samples from $P_{\theta}(\mathbf{x_{mis}}|\mathbf{z}) Q_{\phi}(\mathbf{z}|\mathbf{x_{obs}})$ obtained by ancestral sampling. More specifically, we get samples from $Q_{\phi}(\mathbf{z}|\mathbf{x_{obs}})$ by marginalizing out $\mathbf{r}$ in $Q_{\phi}(\mathbf{z}|\mathbf{r}, \mathbf{x_{obs}}, \mathbf{m})$ and $Q_{\phi}(\mathbf{x_{mis}}| \mathbf{z}, \mathbf{r}, \mathbf{x_{obs}}, \mathbf{m})$. The importance weights are then defined by
\begin{align*}
    w^{(l)} = \sum_{\mathbf{r}=1}^k  \frac{P_{\theta}(\mathbf{x_{mis}}^{(l)}|\mathbf{z}^{(l)}, \mathbf{r}) P_{\theta}(\mathbf{z}^{(l)}|\mathbf{r})P_{\theta}(\mathbf{r})}{  Q_{\phi}({\mathbf{z}}^{(l)}|\mathbf{r}, {\mathbf{x_{obs}}})Q_{\phi}(\mathbf{r}|\mathbf{x_{obs}}, \mathbf{m})}.
\end{align*}
Note that, in contrast to \cite{mat18}, we do not rely on such an IWAE-based approach in the training stage. This stems from the fact that we propose a complex variational distribution and thus do not need the additional complexity induced by an importance-weighted approach \citep{cremer2017reinterpreting, mat18}. This hypothesis is further supported by our results in the Experiments section. 

While single imputation returns a single estimator for the missing data, it is usually more interesting to sample multiple values from the predictive distribution for uncertainty quantification and statistically robust inference in downstream tasks \citep{rubin1996multiple}. Multiple imputations can be obtained by sampling multiple times from the generative model. Again we follow \cite{mat18} and use sampling importance resampling. For this, we generate a set of imputations and weight them using the weights defined above. The multiple imputations are then sampled from this weighted set.
\begin{table*}[ht!]
\centering
\caption{RMSE (Average$\pm$Std of RMSE) for single imputation on data with a missingness rate of 20\%.}
\label{20}
\begin{tabularx}{\textwidth}{|X||XX|XX|XX|}\hline
Model class & \multicolumn{2}{c}{Adult} \vline & \multicolumn{2}{c}{Letter} \vline & \multicolumn{2}{c}{Wine} \vline \\ \hline 
Missingness & MCAR & MNAR & MCAR & MNAR & MCAR & MNAR\\ \hline 
 PSMVAE(a) & .2494$\pm${.0021} & .4943$\pm$.3187 & .0964$\pm$.0013 & \textbf{.0835}$\pm$\textbf{.0153} & .0958$\pm$.0054 & \textbf{.1034}$\pm$\textbf{.0026}\\
PSMVAE(b) & {.2426}$\pm${.0019} & .5249$\pm$.2387 & .0936$\pm$.0008 & .0864$\pm$.0152 & .0890$\pm$.0029 & .1158$\pm$.0095\\
\rotatebox[origin=c]{180}{$\Lsh$} K=10.000 & \textbf{.2306}$\pm$\textbf{.0019} & .4981$\pm$.2176 & \textbf{.0879$\pm$.0006} & .0854$\pm$.0162 &  \textbf{.0832}$\pm$\textbf{.0022} & .1069$\pm$.0100 \\
\rotatebox[origin=c]{180}{$\Lsh$} w/o M & .2450$\pm$0.031 & \textbf{.4815}$\pm$\textbf{.2778} & .0941$\pm$.0009 & .0908$\pm$.0185  & .0885$\pm$.0024 & .1167$\pm$.0096\\ 
DLGM & .2467$\pm$.0033 & .5468$\pm$.2328 & .0947$\pm$.0172 & .0947$\pm$.0172 & .0923$\pm$.0036 & .1234$\pm$.0104\\
HIVAE & .2693$\pm$.0338 & .4907$\pm$.2072 & .1023$\pm$.0008 & .0947$\pm$.0179& .0940$\pm$.0032 & .1246$\pm$.0156\\
VAE & .2562$\pm$.0027 & .5012$\pm$.2313 & .1119$\pm$.0008 & .1061$\pm$.0194  & .1067$\pm$.0035 & .1255$\pm$.0105\\ \hline
MIWAE  & .2845$\pm$.0121 & .6081$\pm$.2423 & .1183$\pm$.0018 & \textbf{.1024}$\pm$\textbf{.0191} & .1129$\pm$.0034 & .1253$\pm$.0259\\
\rotatebox[origin=c]{180}{$\Lsh$} K=10.000 & \textbf{.2373}$\pm$\textbf{.0015} & .5872$\pm$.3065 & \textbf{.1149}$\pm$\textbf{.0004} & .1242$\pm$.0063 & \textbf{.0915}$\pm$\textbf{.0017} & {.0803}$\pm${.0117}\\
not-MIWAE & .2374$\pm$.0011 & \textbf{.5201}$\pm$\textbf{.2640} & .1153$\pm$.0005  & .1192$\pm$.0317 & .0928$\pm$.0022 & \textbf{.0756}$\pm$\textbf{.0089}\\ 
GAIN & .2570$\pm$.0084 & .5940$\pm$.3744 & .1518$\pm$.0074 & .1316$\pm$.0061 & .1749$\pm$.0042 & .1151$\pm$.0175\\ \hline
MICE & .2383$\pm$.0013 & .5879$\pm$.3079 & .1167$\pm$.0008 & .1235$\pm$.0069  & .0881$\pm$.0030 & .0782$\pm$.0117\\
\rotatebox[origin=c]{180}{$\Lsh$} sample & .3166$\pm$.0015 & .6100$\pm$.2375 & .1575$\pm$.0007 & .1664$\pm$.0128 & .1264$\pm$.0025 & .1073$\pm$.0135\\
MissForest & \textbf{.2246}$\pm$\textbf{.0026} & \textbf{.4513}$\pm$.\textbf{1774} & \textbf{.0650}$\pm$\textbf{.0018} & \textbf{.0534}$\pm$\textbf{.0113} & \textbf{.0738}$\pm$\textbf{.0023} & \textbf{.0698}$\pm$\textbf{.0035}\\
mean & .2510$\pm$.0012 & .6676$\pm$.3350 & .1560$\pm$.0005 & .1938$\pm$.0341  & .1765$\pm$.0041 & .1591$\pm$.0292\\
\hline
\end{tabularx}
\end{table*}

\section{EXPERIMENTS}
In this section, we quantitatively evaluate the imputation performance of our model (from now on PSMVAE) and several state-of-the-art imputation approaches on UCI data sets \citep{lichman2013uci} commonly used in the machine learning for imputation literature \citep{yoo18, nazabal2020handling}. 
Each experiment is repeated five times with different seeds. We report the root mean squared error (RMSE) of the estimators of the missing data along with its standard deviations across the experiments. Unless otherwise specified, we choose $\pi'=0.5$. We further use the same hyperparameter constellation for each data set. Please see the supplementary material for details, the complete results and additional experiments such as the multiple imputation setting. The complete code for all experiments can be found online and figures are created using Weights \& Biases \citep{wandb}.

We synthetically introduce missingness into our data sets. The MNAR data are generated by self-masking one of the features. Every time this randomly selected variable is larger than its median we set it to zero with probability equal to the missingness rate, similar to \cite{ipsen2020not}. The MCAR characteristic is introduced by randomly setting a feature to be missing with probability equal to the missingness rate. We present an overview of some results in Table~\ref{20} and Table~\ref{80}. Note that the best score of each group is highlighted in bold, but that some of the approximate confidence intervals overlap.

\subsection{Properties Of The Model}
First, we analyze how the performance of our model changes due to incremental modifications in its architecture. For this purpose we compare a basic VAE, a simplified HIVAE model (HIVAE without different losses for different categories), and a deep latent Gaussian Variable model (DLGM, which is a VAE with one categorical and two Gaussian latent variables) with our model. We train our algorithm once with one importance sample and no missingness mask as input and output (PSMVAE w/o M), once with one importance sample and generative model as specified in Figure~\ref{fig:pmz} (PSMVAE(b)), once as PSMVAE(b) with 10,000 importance samples, and finally with generative model as specified in Figure~\ref{fig:pmxr} (PSMVAE(a)) and 10,000 importance samples.

We note in our experiments that adding a categorical variable to a basic VAE always leads to a reduction in the RMSE loss of up to $1.4$ percentage points. Moreover, we find that the PSMVAE w/o M is performing better than the DLGM. The main difference between the models is the assumed probabilistic semi-supervision of $\mathbf{x_{mis}}$ in the PSMVAE(b) while $\pi'$ is equal to 0 in the DLGM.\footnote{Please refer to the supplementary material for a detailed description of the DLGM.} The average improvement of approximately $4.78$\% compared to the DLGM loss across all data sets speaks for the semi-supervised approach. Our model is also consistently performing better than the HIVAE which is a special case of our model when the union of $\mathbf{x_{mis}}$ and $\mathbf{x_{obs}}$ are assumed to be $d$-dimensional and $\pi'=0$. In four of the displayed twelve experiments, from which three were run on data with MNAR, including the missingness mask led to a higher loss. We hypothesize that this stems from the class imbalance: While only one of the variables is set to be MNAR, the other variables are always observed.

\begin{table*}[!ht]
\centering 
\caption{Pattern-set accuracy on data with a missingness rate of 20\%.}\label{tab:clusteracc}
\resizebox{\linewidth}{!}{%
\begin{tabular}{|l||ll|ll|ll|}\hline
 & \multicolumn{2}{c}{Breast} \vline & \multicolumn{2}{c}{Wine} \vline & \multicolumn{2}{c}{Spam} \vline  \\ \hline
Algorithm & \tiny{MNAR} & \tiny{MNAR+MCAR} & \tiny{MNAR} & \tiny{MNAR+MCAR} &\tiny{MNAR} & \tiny{MNAR+MCAR}\\ \hline
PSMVAE (a)& \textbf{.7819}$\pm$\textbf{.0937}  & \textbf{.7116}$\pm$\textbf{.1110} & .6113$\pm$.0564 & .6435$\pm$.0659 &.7349$\pm$.0638 & .8137$\pm$.0965 \\
PSMVAE (b) & .7709$\pm$.0824 & .6975$\pm$.1104 & .6068$\pm$.0535 & .6476$\pm$.0843 &\textbf{.7375}$\pm$\textbf{.0622} & .8123$\pm$.1019 \\
HIVAE w.M & .7050$\pm$.1161 &  .6887$\pm$.1079 & \textbf{.6544}$\pm$\textbf{.0813} & \textbf{.7267}$\pm$\textbf{.0725} & .7244$\pm$.0833 & \textbf{.8145}$\pm$\textbf{.0949}  \\
\hline
\end{tabular}%
}
\end{table*}

In Table \ref{tab:clusteracc}, we see that our models learn to cluster the data into meaningful missingness pattern-sets. For this purpose we ran our models and a HIVAE, whose input and output was a concatenation of the observed data and the corresponding missingness masks, on a variety of data sets. We did not only model MNAR, but also created a version of the data sets where all variables were MCAR and one variable was additionally MNAR. We assume that the model learns to cluster the data if the categories of the (here two-dimensional) latent variable $\mathbf{r}$ correspond to the cases when the variable that is MNAR is higher resp. lower than its median, which is the threshold for the assignment to the different missingness mechanisms. The pattern-set accuracy was then determined by computing the accuracy of a permutation $\widetilde{p}$ of the learned categories $\mathbf{r}|\mathbf{x_{obs}}, \mathbf{m}$ where the permutation was chosen such that $\widetilde{p}(\mathbf{r}|\mathbf{x_{obs}}, \mathbf{m})=0$ contained the highest fraction of observations of the MNAR pattern-set. Even though we only induce 20\% missingness we see that all models learn to cluster the data.

\begin{table*}[!ht]
\centering
\caption{RMSE (Average$\pm$Std of RMSE) for single imputation on data with a missingness rate of 80\%.}
\label{80}
\begin{tabularx}{\textwidth}{|X||XX|XX|XX|}\hline
 & \multicolumn{2}{c}{Breast} \vline & \multicolumn{2}{c}{Credit} \vline & \multicolumn{2}{c}{Spam} \vline  \\ \hline
Algorithm & MCAR & MNAR & MCAR & MNAR & MCAR & MNAR\\ \hline 
 PSMVAE(a) & \textbf{.1003}$\pm$\textbf{.0035} & .1114$\pm$.0417 & .1715$\pm$.0023 & \textbf{.0948}$\pm$\textbf{.061} &  .0603$\pm$.0022 & .0801$\pm$.0232\\
 PSMVAE(b) & {.1004}$\pm${.0013}  & .1085$\pm$.0258 & .1712$\pm$.0010 & {.1095}$\pm${.0533}  & {.0591}$\pm${.0023} & .0889$\pm$.0427\\
\rotatebox[origin=c]{180}{$\Lsh$} K=10,000 & .1125$\pm$.0034 & \textbf{.0729}$\pm$\textbf{.0184} & \textbf{.1701}$\pm$\textbf{.0011} & {.0973}$\pm${.0437} & 
\textbf{.0552}$\pm$\textbf{.0035} & \textbf{.0800}$\pm$\textbf{.0377}\\
 \rotatebox[origin=c]{180}{$\Lsh$}	 w/o M & .1058$\pm$.0025 & .1025$\pm$.0276 & .1730$\pm$.0033 & .1202$\pm$.0555 & .0594$\pm$.0023 & .0879$\pm$.0433\\ 
 DLGM & .1179$\pm$.0011 & .1303$\pm$.0363 & .1742$\pm$.0026 & .1458$\pm$.0507 & .0624$\pm$.0027 & .0943$\pm$.0415\\ 
 HIVAE & .1207$\pm$.0023& .1406$\pm$.0537 & .1743$\pm$.11&.1301$\pm$.0542 & .0621$\pm$.0020 & .0924$\pm$.0426\\
 VAE & .1214$\pm$.0013 &.1214$\pm$.0475 & .1748$\pm$.0011 & .1267$\pm$.0576 & .0614$\pm$.0023 & .0929$\pm$.0425\\ \hline
 MIWAE & .1926$\pm$.0256& .1452$\pm$.0316 & .1807$\pm$.0009 & .1716$\pm$.1262 & .0690$\pm$.0025 & .1133$\pm$.0709\\
\rotatebox[origin=c]{180}{$\Lsh$} K=10,000 & \textbf{.0953}$\pm$\textbf{.0020} & \textbf{.1229}$\pm$\textbf{.0381} & \textbf{.1582}$\pm$\textbf{.0004}  & .1076$\pm$.0391 & \textbf{.0579}$\pm$\textbf{.0021} &  \textbf{.0858}$\pm$\textbf{.0364}\\ 
not-MIWAE & .2795$\pm$.0496 & .1393$\pm$.0425  & .2744$\pm$.0045 & \textbf{.1027}$\pm$\textbf{.0293} & .1112$\pm$.0035 &  .0875$\pm$.0456\\ 
 GAIN & .6115$\pm$.1014 & .9170$\pm$.7307 & .1654$\pm$.0024 & .1334$\pm$.0649 & .0603$\pm$.0020 & .0886$\pm$.0423  \\ \hline
MICE  & .1343$\pm$.0044 & .1445$\pm$.0424& .1671$\pm$.0036 & \textbf{.1113}$\pm$\textbf{.0356} & .0619$\pm$.0023 & \textbf{.0886}$\pm$\textbf{.0372}\\
\rotatebox[origin=c]{180}{$\Lsh$} sample & .1419$\pm$.0021 & .1526$\pm$.0392 & .1961$\pm$.0010 & .1664$\pm$.0128 & .0804$\pm$.0027 & .1000$\pm$.0418\\
 MissForest & \textbf{.1155}$\pm$\textbf{.0020}  & \textbf{.1260}$\pm$\textbf{.0269}& .1759$\pm$.0028 & .1145$\pm$.0384 & .0732$\pm$.0026 & .0909$\pm$.0397\\
 Mean & .1496$\pm$.0010 & .1545$\pm$.0297 & \textbf{.1640}$\pm$\textbf{.0005} & .1666$\pm$.0976 & \textbf{.0600}$\pm$\textbf{.0020} & .0926$\pm$.046\\
\hline
\end{tabularx}
\label{cluster}
\end{table*}
\subsection{Comparison With Other Imputation Approaches}
We compare our proposed method with common imputation methods from the deep learning and statistical literature. 
As the losses of MIWAE and PSMVAE(b) indicate, importance sampling is essential in the imputation step of VAEs to reduce the noise of the imputations.  We further note that traditional methods, such as MissForest and MICE, often outperform neural network based methods. We claim that this roots from the fact that single imputations in linear models, such as MICE learned with Bayesian Ridge, rely on consistent estimators of the expectation over the missing data distributions. When we sample a single value from the predictive distribution of MICE instead, we note that the single sample imputation of our method outperforms MICE. In Table~\ref{80} we see that the deep generative models perform better than MICE and MissForest on multiple data sets when the missingness rate is high. Another drawback of MICE and MissForest towards deep generative methods is that they are typically not scalable to high-dimensional data sets while our proposed models are. 

\begin{figure}[ht!]
    \centering
    \includegraphics[width=0.4\textwidth]{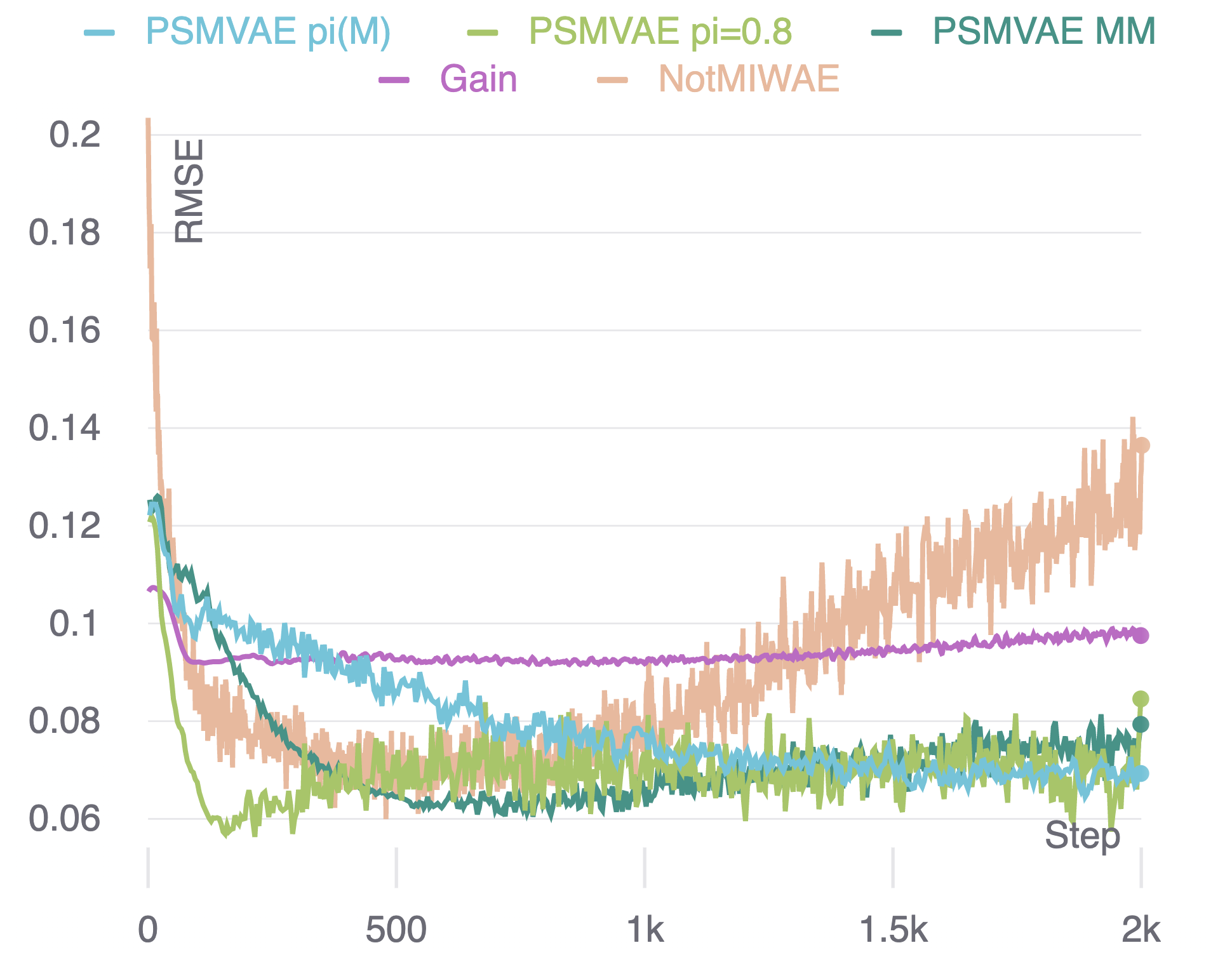}
    \caption{RMSE for single imputation (K=10,000) of various models as a function of iteration steps trained on the Breast data set with 80\% induced MNAR missingness.}
    \label{breastMNAR}
\end{figure}

We further note that our model achieves state-of-the-art imputation performance on both MCAR and MNAR data while MIWAE and Not-MIWAE typically only outperform other methods on one of the missingness types and are thus less robust to the misspecification of the missingness mechanism. In Figure~\ref{breastMNAR} we see that the PSMVAE model with $\pi'(m_j)=\frac{P_{\theta}(m_j=1)}{1+P_{\theta}(m_j=1)}$ (blue line) is the only method with a decreasing imputation loss the longer the method learns. This choice ensures that the distributions of ${x_{obs,j}}$ and ${x_{mis,j}}$ are closer whenever $x_j$ was more likely to be missing and that ${x_{mis,j}}$ is always observed with a probability of at least 50\%. Choosing a small constant $\pi'$ leads to high volatility in the results.  Note that we choose a misspecified missing model (learned by self-masked logistic regression) for training PSMVAE(a) and not-MIWAE. While the loss of both models starts to increase after 500 steps, the loss of PSMVAE(a) increases less steeply than the the loss of not-MIWAE.


\section{CONCLUSION AND FUTURE WORK}
We propose a generative model combining ideas from VAEs and pattern-set mixture models for missing data imputation. This architecture specifies a variational autoencoder that prevents overfitting to data by introducing the assumption of probabilistic semi-supervision of the missing data. We also allow for the specification of a missing model. We see in various experiments with real-world data sets that our model achieves state-oft-the-art imputation performance for different missingness types and explicitly models meaningful clusters which add to the interpretability of the model when no information on the missingness mechanism is available. We recommend our model for use when the underlying data mechanisms are unknown, the missingness is high and/or the data is high-dimensional. Future research will focus on incorporating assumptions on the differences between the missing models of distinct missingness pattern-sets.

\subsubsection*{Acknowledgments}
We thank the reviewers for their constructive feedback. SG is a student of the EPSRC CDT in Modern Statistics and Statistical Machine Learning (EP/S023151/1) and receives funding from the Oxford Radcliffe Scholarship and Novartis. RC is supported by the Engineering and Physical Sciences Research Council (EPSRC) through the Bayes4Health programme Grant EP/R018561/1. LJK is supported by the French government under management of Agence Nationale de la Recherche as part of the ``ABSint'' programme, reference ANR-18-CE40-0034. CH is supported by The Alan Turing Institute, Health Data Research UK, the Medical Research Council UK, the EPSRC through the Bayes4Health programme Grant EP/R018561/1, and AI for Science and Government UK Research and Innovation (UKRI).

\bibliography{sample_paper.bib}

\end{document}


%
\runningtitle{Deep Generative Pattern-Set Mixture Models:
Supplementary {M}aterials}

%
\runningauthor{Sahra Ghalebikesabi, Rob Cornish, Luke J. Kelly,  Chris Holmes}

\onecolumn

\aistatstitle{Deep Generative Pattern-Set Mixture Models\\for Nonignorable Missingness:\\
Supplementary {M}aterials}

\aistatsauthor{ Sahra Ghalebikesabi \And Rob Cornish \And  Luke J. Kelly \And  Chris Holmes}

\aistatsaddress{University of Oxford \And University of Oxford \And CEREMADE, CNRS \\ Université Paris-Dauphine \And University of Oxford} 

\footnotetext[1]{denotes senior author}
\section{PROOFS}
\begin{proof}[Proof of Proposition 1]
Let $\mu$ denote the posterior distributional parameter of $(\mathbf{x_{mis}}\mid\mathbf{x_{obs}}, \mathcal{M})\sim\mathcal{N}(\mu, c)$
for some fixed constant $c$. In the following, we assume that the distributional parameter is found by optimization of the expected likelihood given the augmented dataset
\begin{align}
    \mathbf{E}_{\mathbf{y}}[P_{\theta}(\mathbf{x_{obs}}, \mathbf{x_{mis, 1-y}}, \mathbf{m}\mid\mathcal{D}_{\bm{\pi}}(\mathbf{y}))] = \bm{\pi}P_{\theta}(\mathbf{x_{obs}}, \mathbf{m}\mid\mathcal{D}_{\bm{\pi}}(\mathbf{1})) + (1-\bm{\pi})P_{\theta}(\mathbf{x_{obs}}, \mathbf{m}, \mathbf{x_{mis}}\mid\mathcal{D}_{\bm{\pi}}(\mathbf{0})). \label{lik}
\end{align}
Let $\theta$ denote the parameter set of the generative model. We write $\theta_1$ for the parameter set where $\mu=\mu_1$ and $\theta_2$ for the parameter set where $\mu=\mu_2$.
For any two parameters $\mu_1$ and $\mu_2$ with
\begin{align*}
\bm{\pi}P_{\theta_1}(\mathbf{x_{obs}}, \mathbf{m}\mid\mathcal{D}_{\bm{\pi}}(\mathbf{1})) + (1-\bm{\pi})P_{\theta_1}(\mathbf{x_{obs}}, \mathbf{m}, \mathbf{x_{mis}}\mid\mathcal{D}_{\bm{\pi}}(\mathbf{0})) = &\bm{\pi}P_{\theta_2}(\mathbf{x_{obs}}, \mathbf{m}\mid\mathcal{D}_{\bm{\pi}}(\mathbf{1})) \\
&\quad + (1-\bm{\pi})P_{\theta_2}(\mathbf{x_{obs}}, \mathbf{m}, \mathbf{x_{mis}}\mid\mathcal{D}_{\bm{\pi}}(\mathbf{0})),
\end{align*}
we have to show that $\mu_1=\mu_2$.

From the identifiability of the mean parameter of Gaussian distributions as sample mean in maximum likelihood estimation, it follows that 
\begin{align}
\bm{\pi}P_{\theta_1}(\mathbf{x_{obs}}, \mathbf{m}\mid\mathcal{D}_{\bm{\pi}}(\mathbf{1})) &= \bm{\pi}P_{\theta_2}(\mathbf{x_{obs}}, \mathbf{m}\mid\mathcal{D}_{\bm{\pi}}(\mathbf{1})) \text{ and thus} \nonumber\\
(1-\bm{\pi})P_{\theta_1}(\mathbf{x_{obs}}, \mathbf{m}, \mathbf{x_{mis}}\mid\mathcal{D}_{\bm{\pi}}(\mathbf{0})) &= (1-\bm{\pi})P_{\theta_2}(\mathbf{x_{obs}}, \mathbf{m}, \mathbf{x_{mis}}\mid\mathcal{D}_{\bm{\pi}}(\mathbf{0})) \label{2},
\end{align}
where only the probabilities in Equation \eqref{2} depend on $\mu$. The maximization of $P_{\theta_2}(\mathbf{x_{obs}}, \mathbf{m}, \mathbf{x_{mis}}\mid\mathcal{D}_{\bm{\pi}}(\mathbf{0}))$ equals a maximum likelihood estimation with a fully observed dataset of $\mathbf{x_{mis}}$. Again, from the identifiability of the mean parameter of Gaussian distributions, it follows that $\mu_1=\mu_2$ from \eqref{2}.
\end{proof}

\begin{proof}[Proof of ELBO in Table 1]
The expected likelihood given the augmented dataset \eqref{lik} is a weighted sum over the likelihood of the observed model $P_{\theta}(\mathbf{x_{obs}}, \mathbf{m}, \mathbf{x_{mis}}\mid\mathcal{D}_{\bm{\pi}}(\mathbf{0}))$ and the likelihood of the unobserved model $P_{\theta}(\mathbf{x_{obs}}, \mathbf{m}\mid\mathcal{D}_{\bm{\pi}}(\mathbf{1}))$. 

The observed model can be learned by maximizing the ELBO which we can compute using Jensen's inequality:
\begin{align*}
    &\log P_{\theta}(\mathbf{x_{obs}}, \mathbf{m}, \mathbf{x_{mis}}\mid\mathcal{D}_{\bm{\pi}}(\mathbf{0})) = \log P_{\theta}(\mathbf{x_{obs}}, \mathbf{m}, \mathbf{x_{mis}}, \mathbf{r}, \mathbf{z}\mid\mathcal{D}_{\bm{\pi}}(\mathbf{0})) - \log P_{\theta}(\mathbf{r}, \mathbf{z}\mid\mathcal{D}_{\bm{\pi}}(\mathbf{0}))\\
    &\quad\geq \mathbb{E}_{Q_{\phi}(\mathbf{r}, \mathbf{z}\mid\mathbf{x_{obs}}, \mathbf{m}, \mathbf{x_{mis}})} \bigg[\log P_{\theta}(\mathbf{x_{obs}}, \mathbf{m}, \mathbf{x_{mis}}, \mathbf{r}, \mathbf{z}\mid\mathbf{x_{obs}}, \mathbf{m}, \mathbf{x_{mis}}) - \log Q_{\phi}(\mathbf{r}, \mathbf{z}\mid\mathbf{x_{obs}}, \mathbf{m}, \mathbf{x_{mis}})\bigg]\\
    &\quad= \mathbb{E}_{Q_{\phi}(\mathbf{r}, \mathbf{z}\mid\mathbf{x_{obs}}, \mathbf{m}, \mathbf{x_{mis}})} \bigg[\log {P_{\theta}(\mathbf{m}\mid \mathbf{r}, \mathbf{z}, \mathbf{x_{obs}}, \mathbf{x_{mis}})} +
    \log P_{\theta}(\mathbf{x_{obs}}\mid \mathbf{r}, \mathbf{z}) +
    \log P_{\theta}(\mathbf{x_{mis}}\mid \mathbf{r}, \mathbf{z}) \\
    &\qquad\qquad\qquad\qquad\qquad\qquad+\log \frac{P_{\theta}(\mathbf{z}\mid\mathbf{r})}{ Q_{\phi}(\mathbf{z}\mid\mathbf{r},\mathbf{x_{obs}}, \mathbf{m}, \mathbf{x_{mis}})}
    +\log \frac{P_{\theta}(\mathbf{r})}{Q_{\phi}(\mathbf{r}\mid \mathbf{x_{obs}}, \mathbf{m}, \mathbf{x_{mis}})}\bigg]\displaybreak\\
    &\quad= \mathbb{E}_{Q_{\phi}(\mathbf{r}, \mathbf{z}\mid\mathbf{x_{obs}}, \mathbf{m})} \bigg[
    \log {P_{\theta}(\mathbf{m}\mid \mathbf{r}, \mathbf{z}, \mathbf{x_{obs}}, \mathbf{x_{mis}})} +
    \log P_{\theta}(\mathbf{x_{obs}}\mid \mathbf{r}, \mathbf{z}) +
    \log P_{\theta}(\mathbf{x_{mis}}\mid \mathbf{r}, \mathbf{z}) \\
    &\qquad\qquad\qquad\qquad\qquad\qquad+
    \log \frac{P_{\theta}(\mathbf{z}\mid\mathbf{r})}{ Q_{\phi}(\mathbf{z}\mid\mathbf{r},\mathbf{x_{obs}}, \mathbf{m})}
    +\log \frac{P_{\theta}(\mathbf{r})}{Q_{\phi}(\mathbf{r}\mid \mathbf{x_{obs}}, \mathbf{m})}\bigg],
\end{align*}
where $\log P_{\theta}(\mathbf{m}\mid \mathbf{r}, \mathbf{z}, \mathbf{x_{obs}}, \mathbf{x_{mis}}) =\log P_{\theta}(\mathbf{m}\mid \mathbf{r}, \mathbf{x_{obs}}, \mathbf{x_{mis}})$ for PSMVAE(a) and $\log P_{\theta}(\mathbf{m}\mid \mathbf{r}, \mathbf{z}, \mathbf{x_{obs}}, \mathbf{x_{mis}}) =\log P_{\theta}(\mathbf{m}\mid \mathbf{r}, \mathbf{z})$ for PSMVAE(b). The last equation follows from the assumption that $Q_{\phi}(\cdot|\mathbf{x_{obs}}, \mathbf{m}, \mathbf{x_{mis}})=Q_{\phi}(\cdot|\mathbf{x_{obs}}, \mathbf{m})$.

Similarly we obtain the lower bound of the unobserved model:
\begin{align*}
    &\log P_{\theta}(\mathbf{x_{obs}}, \mathbf{m}\mid\mathcal{D}_{\bm{\pi}}(\mathbf{1})) = \log P_{\theta}(\mathbf{x_{obs}}, \mathbf{m}, \mathbf{x_{mis}}, \mathbf{r}, \mathbf{z}\mid\mathcal{D}_{\bm{\pi}}(\mathbf{1})) - \log P_{\theta}(\mathbf{r}, \mathbf{z}, \mathbf{x_{mis}}\mid\mathcal{D}_{\bm{\pi}}(\mathbf{1}))\\
    &\quad\geq \mathbb{E}_{Q_{\phi}(\mathbf{r}, \mathbf{z}, \mathbf{x_{mis}}\mid\mathbf{x_{obs}}, \mathbf{m})} \bigg[\log P_{\theta}(\mathbf{x_{obs}}, \mathbf{m}, \mathbf{x_{mis}}, \mathbf{r}, \mathbf{z}\mid\mathbf{x_{obs}}, \mathbf{m}) - \log Q_{\phi}(\mathbf{r}, \mathbf{z}, \mathbf{x_{mis}}\mid\mathbf{x_{obs}}, \mathbf{m})\bigg]\\
    &\quad= \mathbb{E}_{Q_{\phi}(\mathbf{r}, \mathbf{z}, \mathbf{x_{mis}}\mid\mathbf{x_{obs}}, \mathbf{m})} \bigg[
    \log {P_{\theta}(\mathbf{m}\mid \mathbf{r}, \mathbf{z}, \mathbf{x_{obs}}, \mathbf{x_{mis}})} +
    \log P_{\theta}(\mathbf{x_{obs}}\mid \mathbf{r}, \mathbf{z}) +
    \log \frac{P_{\theta}(\mathbf{x_{mis}}\mid \mathbf{r}, \mathbf{z})}{Q_{\phi}(\mathbf{x_{mis}}\mid \mathbf{x_{obs}}, \mathbf{m}, \mathbf{z}, \mathbf{r})}
    \\
    &\qquad\qquad\qquad\qquad\qquad\qquad +
    \log \frac{P_{\theta}(\mathbf{z}\mid\mathbf{r})}{ Q_{\phi}(\mathbf{z}\mid\mathbf{r},\mathbf{x_{obs}}, \mathbf{m}, \mathbf{x_{mis}})}
    +\log \frac{P_{\theta}(\mathbf{r})}{Q_{\phi}(\mathbf{r}\mid \mathbf{x_{obs}}, \mathbf{m}, \mathbf{x_{mis}})}\bigg],\\
\end{align*}
where $\log P_{\theta}(\mathbf{m}\mid \mathbf{r}, \mathbf{z}, \mathbf{x_{obs}}, \mathbf{x_{mis}})$ is defined as before.

The weighted sum of the ELBOs (denoted by $\mathcal{L}(\mathbf{x_{obs}}, \mathbf{m})$ in the following) can then be written as
\begin{align*}
    \mathcal{L}(\mathbf{x_{obs}}, \mathbf{m})
        &= \mathbb{E}_{Q_{\phi}(\mathbf{r}, \mathbf{z}\mid\mathbf{x_{obs}}, \mathbf{m})} \bigg[
    \log P_{\theta}(\mathbf{x_{obs}}\mid \mathbf{r}, \mathbf{z}) +
     \log \frac{P_{\theta}(\mathbf{z}\mid\mathbf{r})}{ Q_{\phi}(\mathbf{z}\mid\mathbf{r},\mathbf{x_{obs}}, \mathbf{m})}
    +\log \frac{P_{\theta}(\mathbf{r})}{Q_{\phi}(\mathbf{r}\mid \mathbf{x_{obs}}, \mathbf{m})
}\bigg] \\
    &\quad+
    \pi \mathbb{E}_{Q_{\phi}(\mathbf{r}, \mathbf{z}, \mathbf{x_{mis}}\mid\mathbf{x_{obs}}, \mathbf{m})} \bigg[ \log {P_{\theta}(\mathbf{m}\mid \mathbf{r}, \mathbf{z}, \mathbf{x_{obs}}, \mathbf{x_{mis}})} + \log \frac{P_{\theta}(\mathbf{x_{mis}}\mid \mathbf{r}, \mathbf{z})}{Q_{\phi}(\mathbf{x_{mis}}\mid \mathbf{x_{obs}}, \mathbf{m}, \mathbf{z}, \mathbf{r})} \bigg] \\
    &\quad+ (1 - \pi) \mathbb{E}_{Q_{\phi}(\mathbf{r}, \mathbf{z}\mid\mathbf{x_{obs}}, \mathbf{m}, \mathbf{x_{mis}})} \bigg[ \log {P_{\theta}(\mathbf{m}\mid \mathbf{r}, \mathbf{z}, \mathbf{x_{obs}}, \mathbf{x_{mis}})} + \log P_{\theta}(\mathbf{x_{mis}}\mid \mathbf{r}, \mathbf{z}) 
  \bigg].
\end{align*}
Specifying $\pi=1$ when $m_j=0$ yields the ELBO, as shown in Table 1 of the main paper.
\end{proof}

\section{DETAILS OF DATASETS AND EXPERIMENTS}
\subsection{Data}
We split the datasets into a train, a validation and a test dataset respectively based on a 8:1:1 data split. The data is then normalized and mean imputed. The datasets can be found online in the GitHub repository. We compute the normalized RMSE which corresponds to the RMSE on the normalized data. We predict categorical and nominal variables by rounding the predictions of the models. Please refer to Table \ref{data}. Note that our results deviate from those of other papers because we do a train test split for the data which results in a smaller dataset in the training stage.
\begin{table}[ht]
    \begin{center}
    \begin{tabular}{|c||c|c|c|} 
     \hline
     Dataset & Sample size & \# continuous features & \# discrete features 
     \\ \hline
     Adult & 32.561 & 3 & 8 
     \\
     Breast & 569 & 30 & 0  
     \\
     Credit & 30.000 & 14 & 9  
     \\
     Letter & 20.000 & 0 & 16 
     \\
     Spam & 4.601 & 57 & 0 
     \\
     Wine & 6.497 & 11 & 1 
     \\
     \hline
    \end{tabular}
    \end{center}
    \caption{Statistics of the datasets} \label{data}
\end{table}

\subsection{Implementation Details}

We use PyTorch to implement the VAE models. The tensorflow code for GAIN can be found online (\url{https://github.com/jsyoon0823/GAIN}) as well as the PyTorch implementation of MIWAE (\url{https://github.com/pamattei/miwae}). The code for not-MIWAE was provided by the main author of the corresponding paper, Niels Bruun Ipsen. We use the {M}issForest implementation of sklearn (IterativeImputer with ExtraTrees{R}egressor as estimator) and the {M}ICE implementation of sklearn (IterativeImputer with Bayesian{R}idge{R}egressor). 

We train all benchmark models with default parameters as specified in our code. We run all deep learning methods for 1,000 epochs and train them using the Adam optimizer \citep{kingma2014adam}. All subgraphs of the VAE-based approaches have one hidden layer with 128 nodes. Unless otherwise stated, we choose the dimension of the latent Gaussian variable $\mathbf{z}$ equal to 20 and the number of categories $k$ of the latent categorical variable $\mathbf{r}$ as 10. Only {M}IWAE has two hidden layers each with 128 hidden units as specified by \cite{mat18}. We use a rectifier as activation function between two layers. We chose a batch size of 200 for the datasets used in the main paper and a batch size of 512 for MNIST. The number of trees used for MissForest is set to 10. We use the self-masking approach of not-MIWAE where $P(\mathbf{m}\mid\mathbf{x})$ is learned by a logistic regression which is independent for each feature.

The benchmark VAE architectures have been modeled as follows: 
\begin{itemize}
    \item \textit{VAE}: a basic VAE with one Gaussian latent variable learned on the observed data (and not on the missingness mask).
    \item \textit{GMVAE}: a VAE with a Gaussian Mixture prior. That is a VAE with one categorical latent variable $\mathbf{r}$ taking values from $\{1,...,k\}$ and one conditionally normal distributed latent variable $\mathbf{z}\mid \mathbf{r}$. It learns the generative model $P_{}(\mathbf{x_{obs}}|\mathbf{m}, \mathbf{z}, \mathbf{r}) = P_{}(\mathbf{x_{obs}}, \mathbf{m}\mid\mathbf{z}, \mathbf{r}) P_{}(\mathbf{z}\mid \mathbf{r})P_{}(\mathbf{r})$. This model framework is similar to the HIVAE model presented in \citep{nazabal2020handling}. Instead of sampling $\mathbf{r}$ from a Gumbel-Softmax distribution \citep{jang2016categorical}, we instead learn $P_{}(\cdot\mid \mathbf{r}=r)$ for each $r\in\{1,...,k\}$ and weight the resulting loss for each category with its posterior probability.  
    \item \textit{DLGM}: a deep latent Gaussian model which extends the GMVAE by a second latent Gaussian variable $\textbf{w}$ with the same dimensionality as $\mathbf{x}$. It learns the generative distribution $P_{}(\mathbf{x_{obs}}, \mathbf{m}, \mathbf{z}, \mathbf{w}, \mathbf{r}) = P_{}(\mathbf{x_{obs}}, \mathbf{m}\mid\mathbf{z}, \mathbf{r}) P_{}(\mathbf{w}\mid \mathbf{z}, \mathbf{r}) P_{}(\mathbf{z}\mid \mathbf{r})P_{}(\mathbf{r})$. The inference model is structured in the same way as the inference model for the PSMVAE where $\mathbf{x_{mis}}$ is replaced by $\mathbf{w}$.
\end{itemize}
We then impute $\mathbf{x_{mis}}$ by sampling from the conditional distribution of $\mathbf{x_{obs}}$.

In contrast to \cite{nazabal2020handling}, we do not use the Gumbel softmax distribution for sampling the categorical variable $\mathbf{r}$ \citep{jang2016categorical}, but instead integrate out $\mathbf{r}$ in the loss function. See \cite{dilokthanakul2016deep} for a similar approach. 
In the implementation, we weight the log-likelihood of the observed data with the inverse of 1-\textit{missingness rate}. Otherwise increasing the missingness will lead to a higher weight of the log-likelihood of the missingness mask compared to the log-likelihood of the observed data. 

\section{ADDITIONAL EXPERIMENTS}
\subsection{Multiple Imputation}
We follow \cite{mat18} and assess the performance of our models in the multiple imputation setting by computing the test accuracy of the predicted target variable when a one layer classification network is trained on the dataset where each observation with missing entries was imputed 20 times. The test set is again made up of 20 multiple imputations. 

\begin{table}
\centering
\begin{tabularx}{\textwidth}{|X||XX|XX|}\hline
Algorithm & 20\% MCAR & 80\% MCAR & 20\% MNAR & 80\% MNAR
\\
\hline
  PSMVAE(a) &  .9482 $\pm$ .0211&  .8466  $\pm$ .0627 &  .9489 $\pm$ .0116 &  .9501 $\pm$ .0101 \\
  PSMVAE(b) &  .9429  $\pm$ .0180 &  .8791  $\pm$ .0511 &  .9577  $\pm$ .0063 &  .9496  $\pm$ .0157 \\
       MICE &  .9545  $\pm$ .0141 &  .8105 $\pm$ .0337 &  .9578  $\pm$ .0096 &  .9578 $\pm$ .0096 \\
\hline
\end{tabularx}    
\caption{Test accuracy of a one layer neural network for the prediction of breast cancer using the breast dataset which was imputed multiple times}
    \label{tab:my_label}
\end{table}

\subsection{Preliminary Results On Image Inpainting}
\begin{figure}[!ht]
    \centering
    \begin{subfigure}[t]{0.4755\textwidth}
        \centering
        \includegraphics[width=\textwidth]{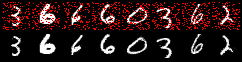}
        \caption{20\% MCAR inpainted using PSMVAE(b)}
    \end{subfigure}%
    ~ 
    \begin{subfigure}[t]{0.4755\textwidth}
        \centering
        \includegraphics[width=\textwidth]{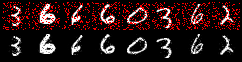}
        \caption{20\% MCAR inpainted using GAIN}
    \end{subfigure}
    
    \begin{subfigure}[t]{0.475\textwidth}
        \centering
        \includegraphics[width=\textwidth]{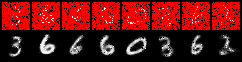}
        \caption{80\% MCAR inpainted using PSMVAE(b)}
    \end{subfigure}%
    ~ 
    \begin{subfigure}[t]{0.475\textwidth}
        \centering
        \includegraphics[width=\textwidth]{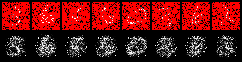}
        \caption{80\% MCAR inpainted using GAIN}
    \end{subfigure}
    \caption{MNIST images with pixels MCAR at different rates inpainted with different imputation algorithms}
    \label{mnist}
\end{figure}

We also show that our method can be used for image inpainting using the MNIST dataset \citep{lecun2010mnist}. We induced missingness completely random as before. 
We then trained our method using 500 epochs, a batch size of 512, $\pi=0$ and only one importance sample. 
Please refer to Figure \ref{mnist} for the results. Missing pixels were highlighted in red in the first row of each subfigure. The second row shows inpainted images using the corresponding imputation algorithm. As we see, the PSMVAE(b) yields sharper images than GAIN does. Especially when the missingness is high, the imputations of our model suffer from considerably less noise than those of GAIN. 

\subsection{Robustness Study}
\begin{figure*}[!ht]
    \centering
    \begin{subfigure}[t]{0.39\textwidth}
        \centering
        \includegraphics[height=2in]{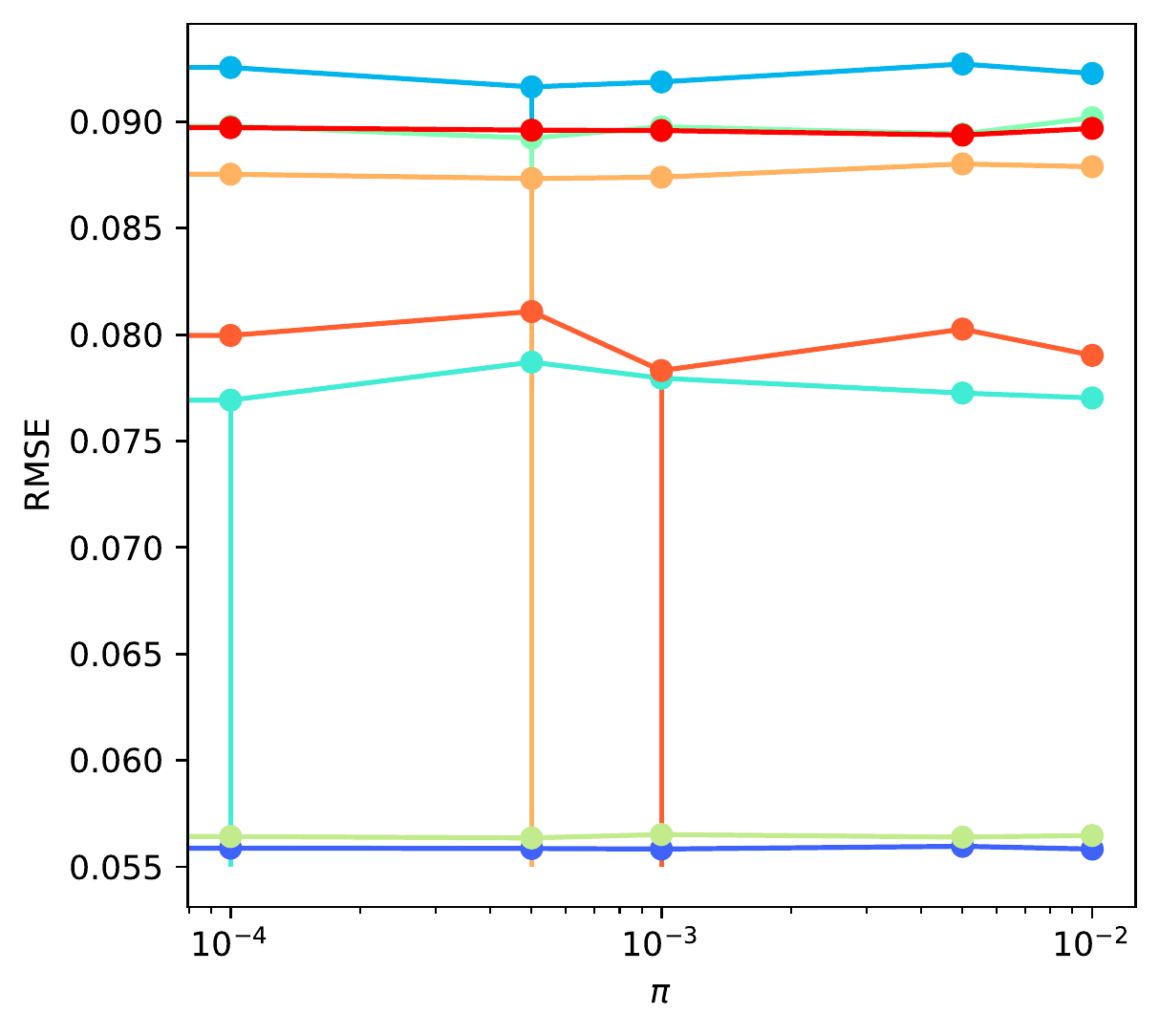}
        \caption{20\% missingness rate}
    \end{subfigure}%
    ~ 
    \begin{subfigure}[t]{0.59\textwidth}
        \centering
        \includegraphics[height=2in]{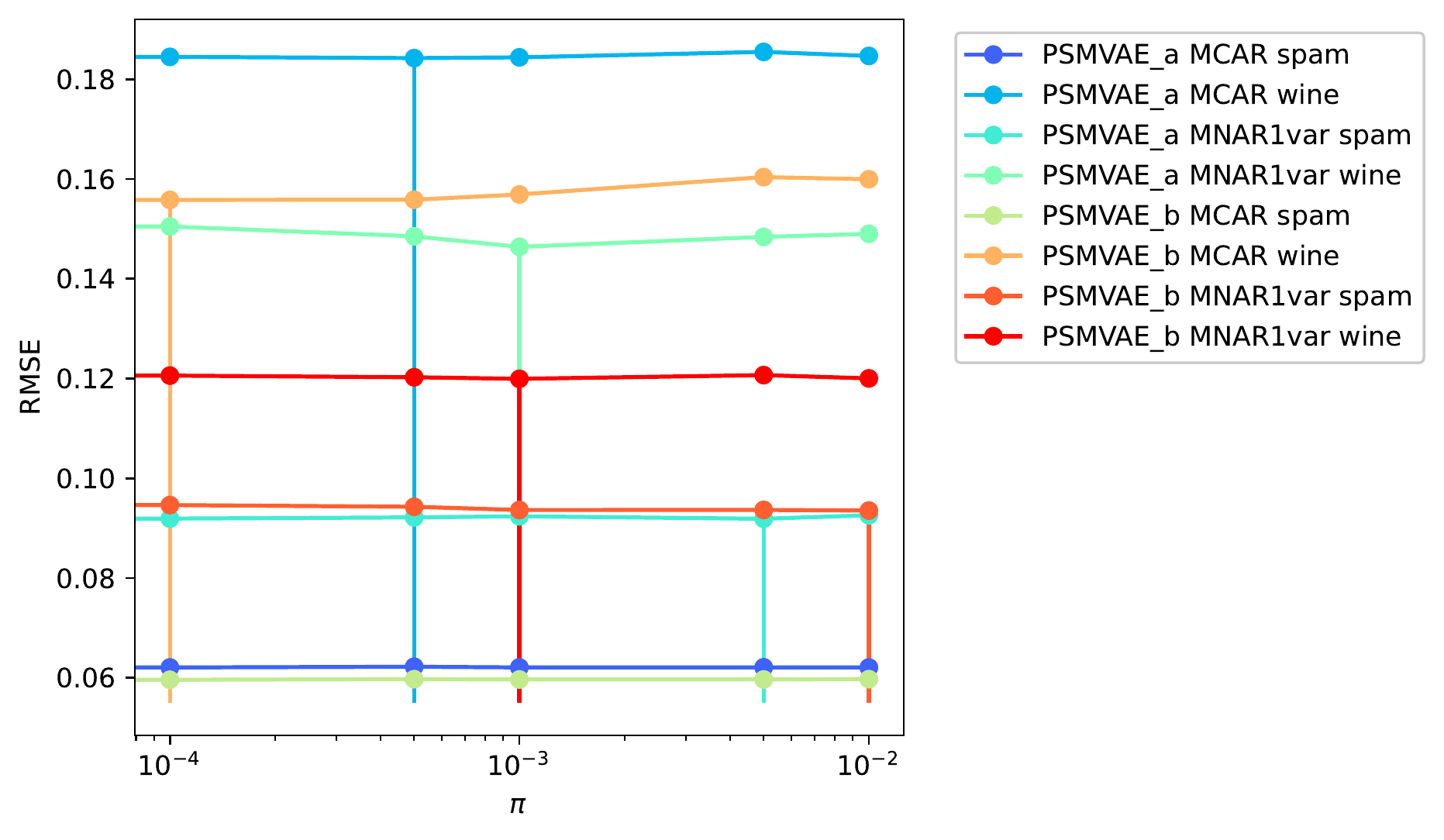}
        \caption{80\% missingness rate \quad\quad\quad\quad\quad\quad\quad}
    \end{subfigure}
    \caption{RMSE of imputation for different values of $\pi$ for a missingness rate of $20\%$ (left) and a missingness rate of $80\%$ (right)}
    \label{pi}
\end{figure*}

We now assess the robustness of our method. Please refer to Figure \ref{pi} for an illustration of the relationship between the RMSE and the hyperparameter $\pi$. The vertical lines highlight the minimum of the loss curves. We notice that the optimal value of $\pi$ is usually on a similar scale as the optimal value of the weight decay parameter. Only on the MCAR spam dataset it is optimal to have a $\pi=0$. Please note that the decay of the curves seems infinitesimal because we compare different methods and datasets within the same figure.

\begin{figure*}[!ht]
    \centering
    \begin{subfigure}[t]{0.39\textwidth}
        \centering
        \includegraphics[height=2in]{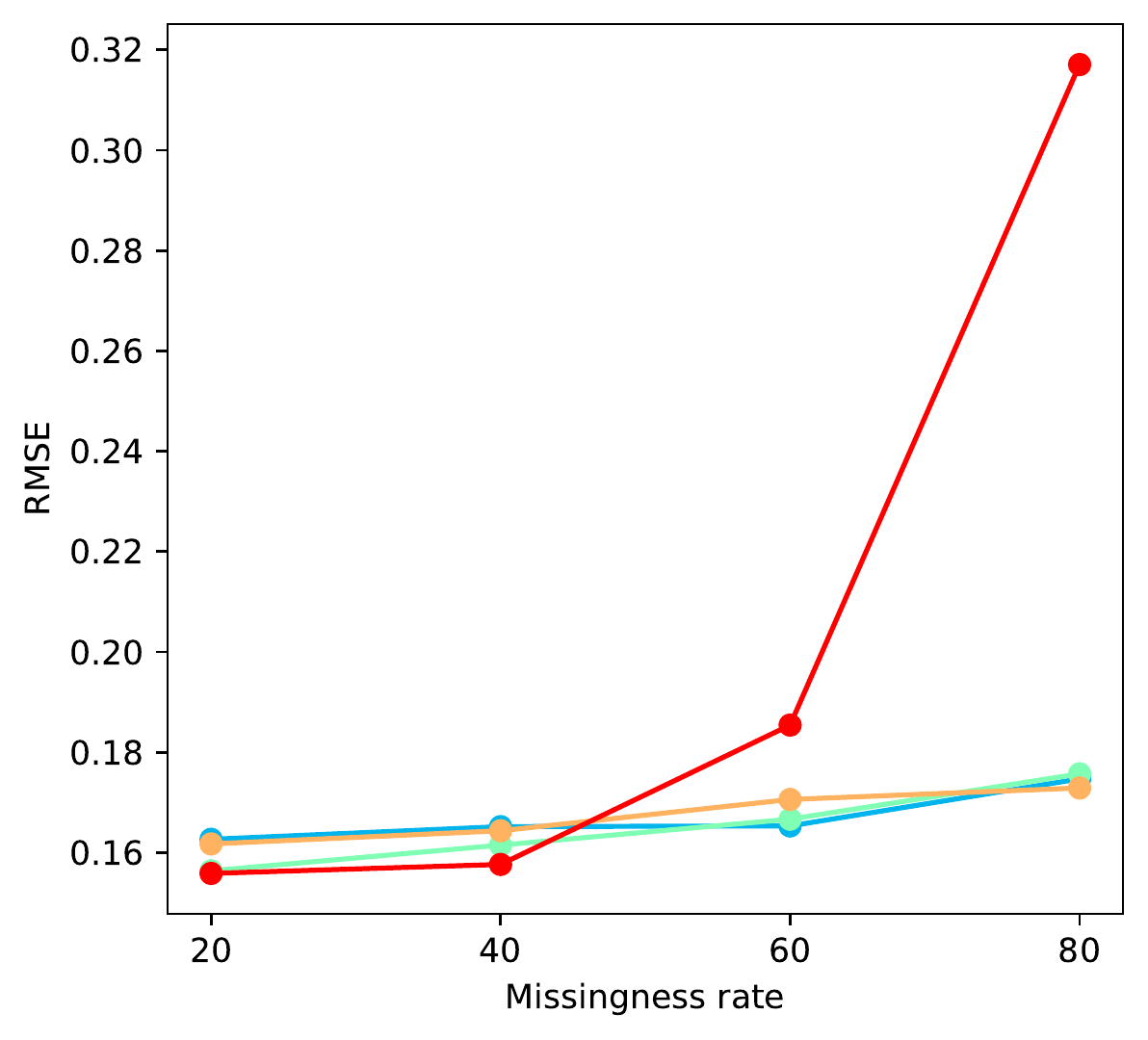}
        \caption{MCAR}
    \end{subfigure}%
    ~ 
    \begin{subfigure}[t]{0.59\textwidth}
        \centering
        \includegraphics[height=2in]{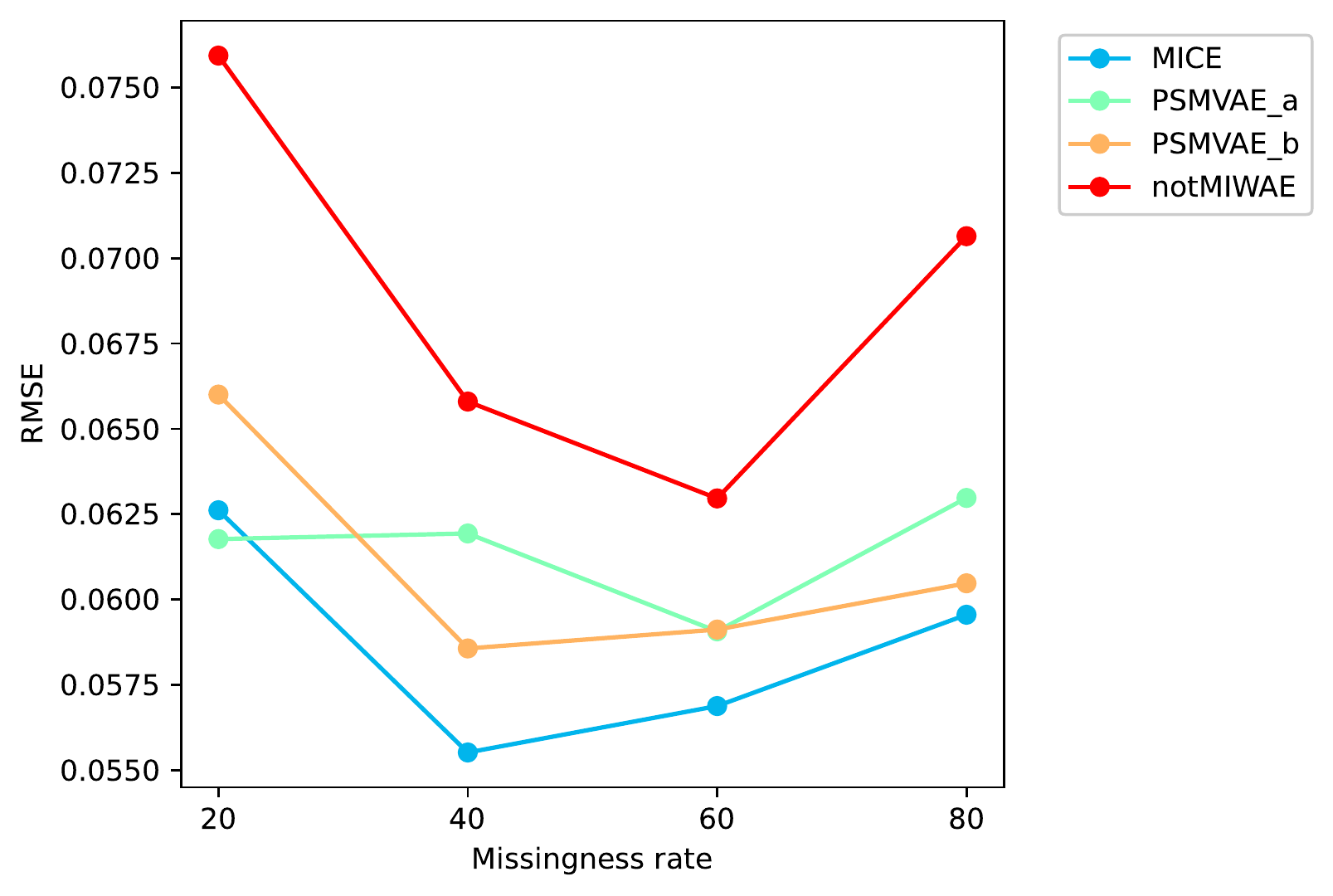}
        \caption{MNAR \quad\quad\quad\quad\quad}
    \end{subfigure}
    \caption{RMSE of imputation for different missingness rates (in \%) when data of the credit dataset are missing}
    \label{rate}
\end{figure*}

In the following we compare the robustness of our models and two benchmarks on the credit dataset. When assessing the relationship of the RMSE and the missingness rate (see Figure \ref{rate}), we note that not-MIWAE is not robust to increasing the missingness rate. This could stem from the fact that increasing the missingness increases the fraction of the optimization loss that comes from the likelihood of the missingness mask compared to the likelihood of the observed data. The larger the missingness, the better do our models compare relative to MICE. 

\begin{figure*}[!ht]
    \centering
    \begin{subfigure}[t]{0.39\textwidth}
        \centering
        \includegraphics[height=2in]{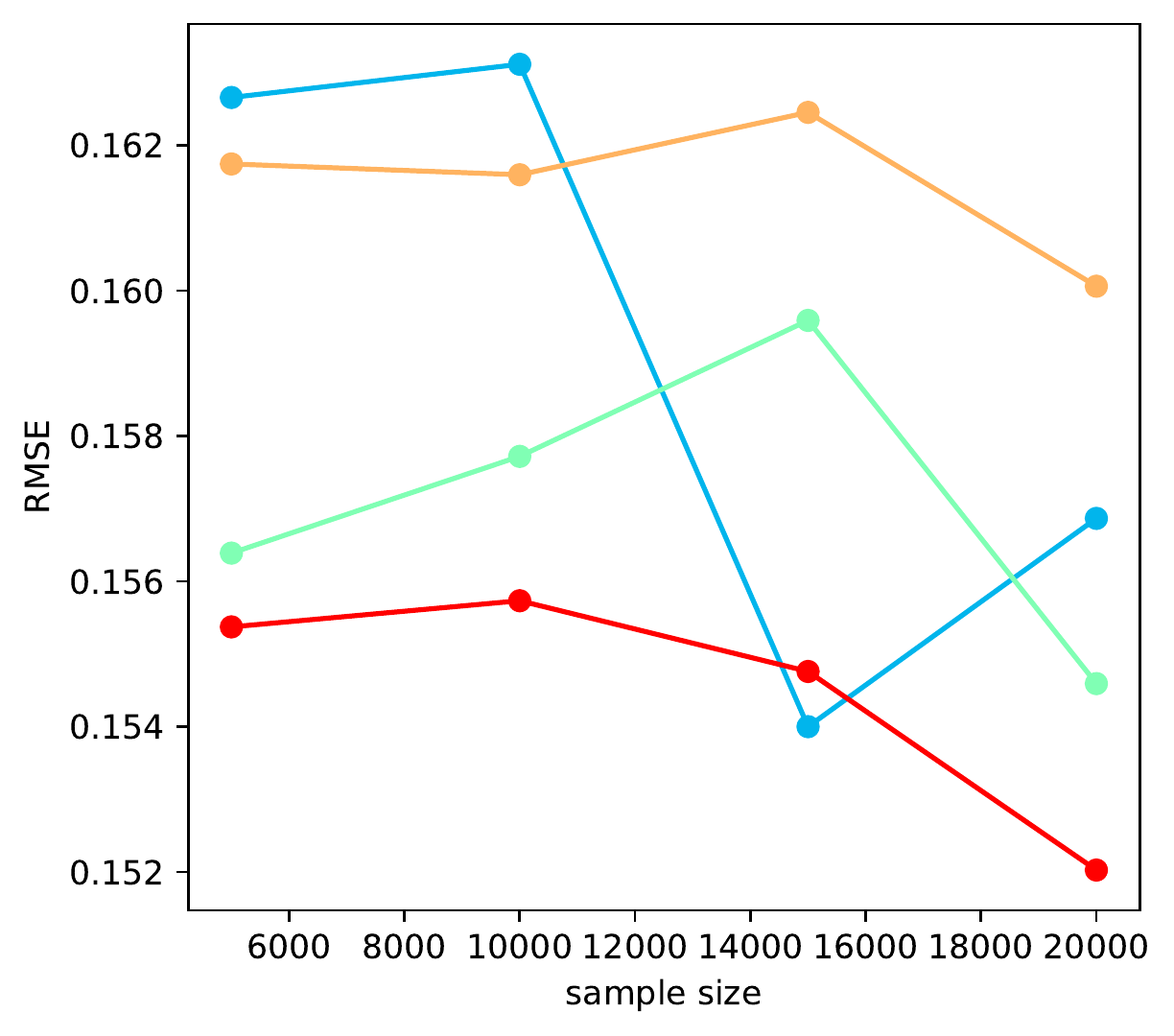}
        \caption{MCAR}
    \end{subfigure}%
    ~ 
    \begin{subfigure}[t]{0.59\textwidth}
        \centering
        \includegraphics[height=2in]{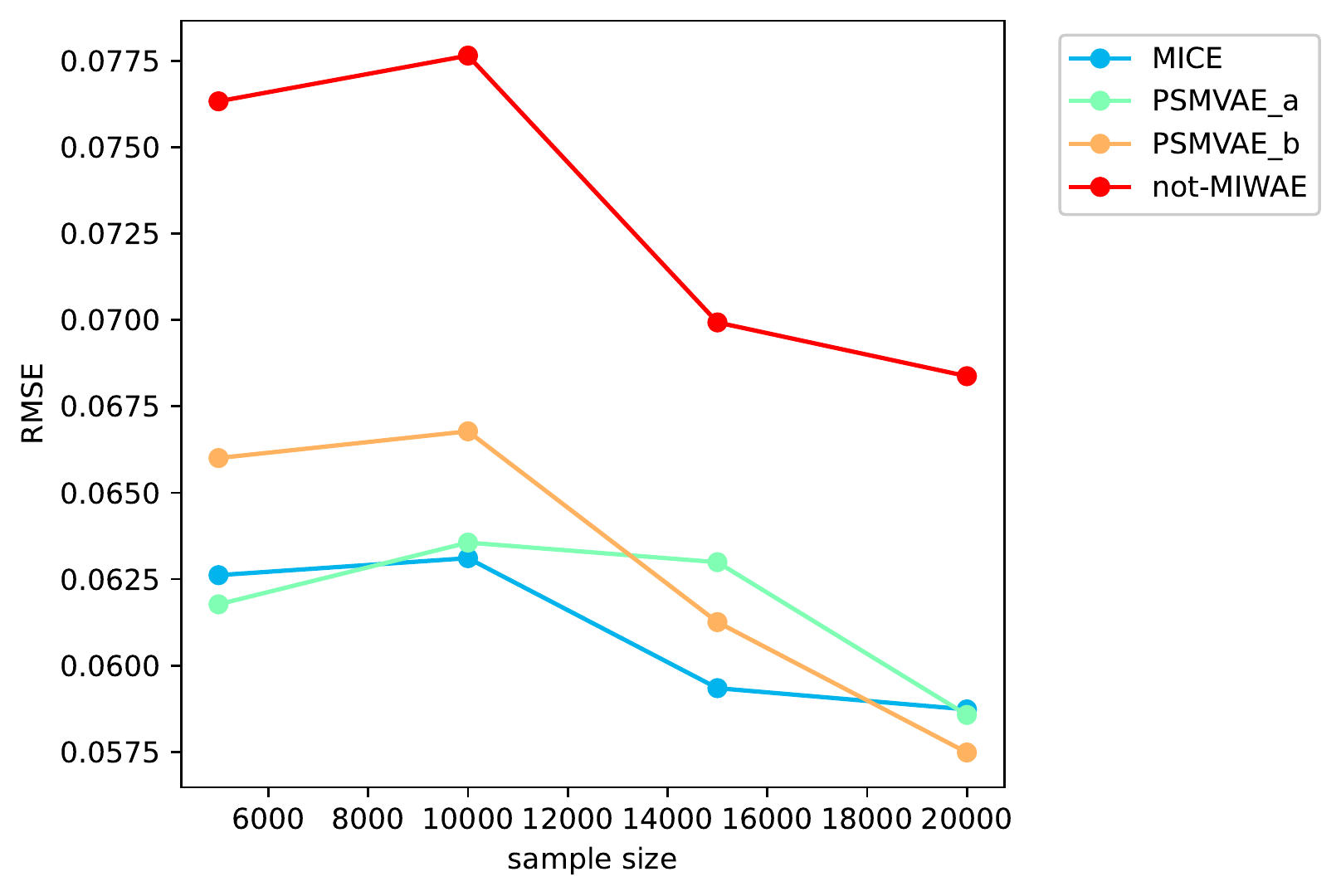}
        \caption{MNAR \quad\quad\quad\quad\quad}
    \end{subfigure}
    \caption{RMSE of imputation for different sample sizes when 20\% of the data of the credit dataset are missing}
    \label{n}
\end{figure*}
\begin{figure*}[!ht]
    \centering
    \begin{subfigure}[t]{0.39\textwidth}
        \centering
        \includegraphics[height=2in]{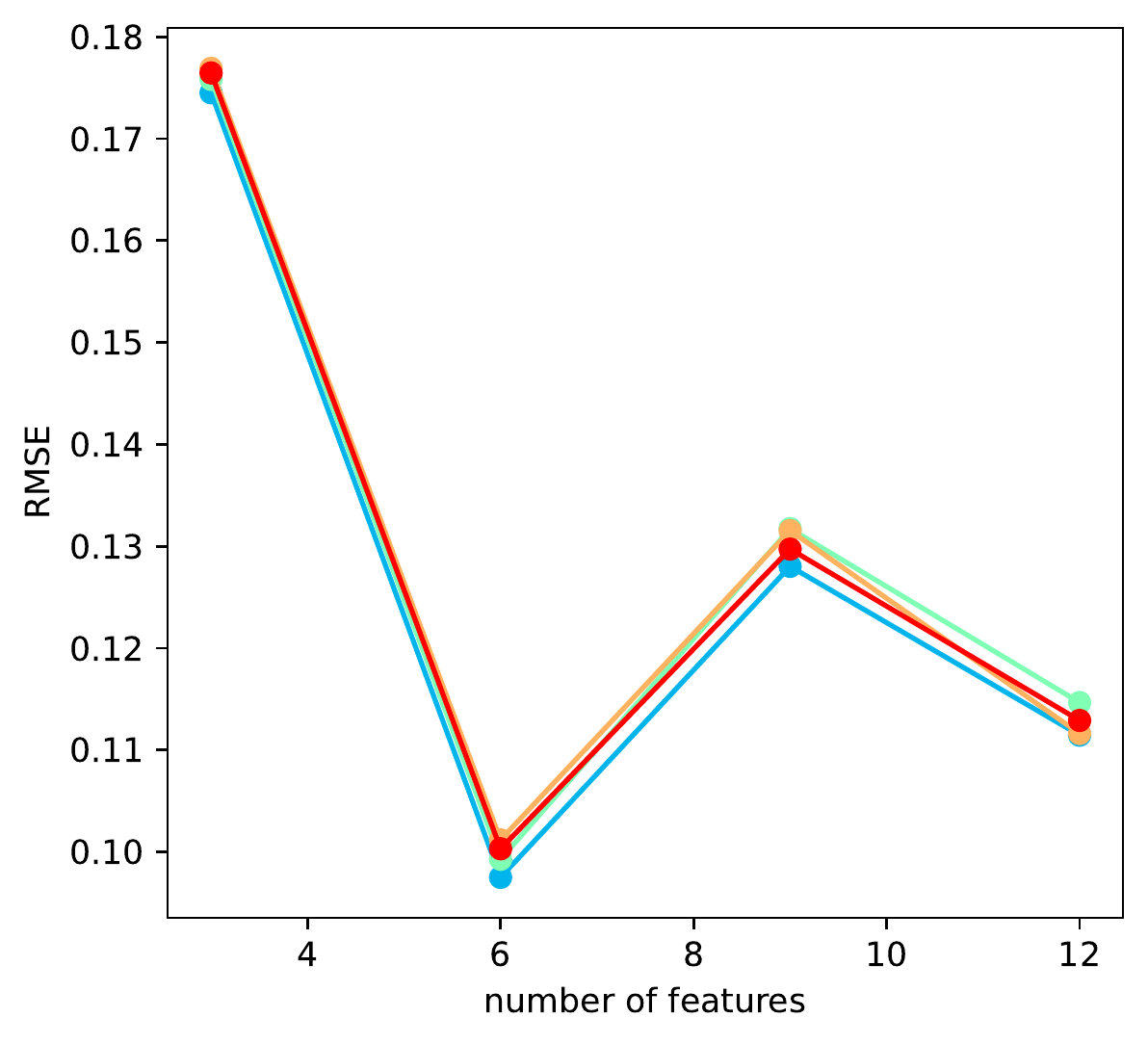}
        \caption{MCAR}
    \end{subfigure}%
    ~ 
    \begin{subfigure}[t]{0.59\textwidth}
        \centering
        \includegraphics[height=2in]{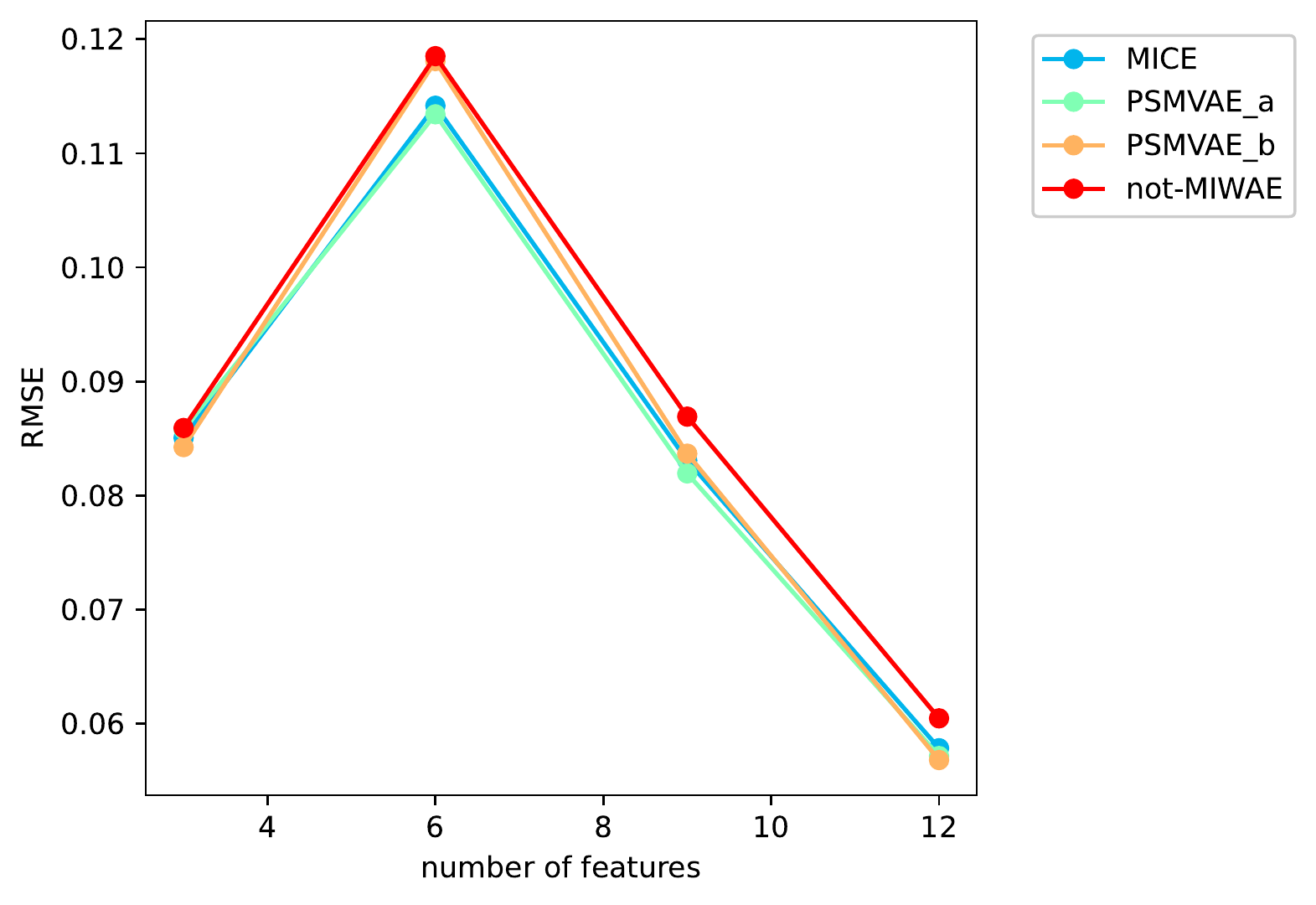}
        \caption{MNAR \quad\quad\quad\quad\quad}
    \end{subfigure}
    \caption{RMSE of imputation for different number of features when 20\% of the data of the credit dataset are missing} \label{m}
\end{figure*}
When we change the sample size of the training dataset (see Figure \ref{n}), we note that the performance of the deep learning methods improves strictly and that the PSMVAE(b) performs better than MICE for large sample sizes.

Eventually, we compare the robustness of the RMSE when the number of features is changed (Figure \ref{m}). We see that the performance of all methods in general decreases when the number of features increases (except for one increase for all methods). The relative performance of our models hereby improves, the more features there are.

\bibliography{supplement}